\documentclass{article}

\usepackage{amsthm}
\usepackage{command_MU}

\usepackage{command_alexis}

\usepackage{command_shortcut}

\usepackage{multirow}
\usepackage{booktabs}
\usepackage{hyperref}
\usepackage{enumitem}
\usepackage{authblk}
\usepackage[normalem]{ulem}

\newcommand{\changer}[1]{{#1}}
\newcommand{\change}[1]{{{#1}}}
\newcommand{\remove}[1]{}

\title{On the Number of Regions of Piecewise Linear Neural Networks\thanks{This work was supported in part by the European Research Council (H2020-ERC Project GlobalBioIm) under Grant 692726  and in part by the Swiss National Science Foundation, Grant 200020\_184646/1.}}

\author[1]{Alexis Goujon \thanks{alexis.goujon@epfl.ch}}

\author[1]{Arian Etemadi}

\author[1]{Michael Unser}

\affil[1]{Biomedical Imaging Group, \'Ecole polytechnique f\'ed\'erale de Lausanne (EPFL)}
\begin{document}
\maketitle


\begin{abstract}
  Many feedforward neural networks \change{(NNs)} generate continuous and piecewise-linear (CPWL) mappings. Specifically, they partition the input domain into regions on which the mapping is affine. The number of these so-called linear regions offers a natural metric to characterize the expressiveness of CPWL \change{NNs}. \change{The} precise determination of this quantity is often out of reach \change{in practice, and} bounds have been proposed for specific architectures, including for ReLU and Maxout \change{NNs}. In this work, \change{we generalize these bounds to NNs with arbitrary and possibly multivariate CPWL activation functions. We first provide upper and lower bounds on the maximal number of linear regions of a CPWL NN given its depth, width, and the number of linear regions of its activation functions.} Our results rely on the combinatorial structure of convex partitions and confirm the distinctive role of depth which, on its own, is able to exponentially increase the number of regions. We then introduce a complementary stochastic framework to estimate the average number of linear regions produced by a \change{CPWL NN}. Under reasonable assumptions, the expected density of linear regions along any 1D path is bounded by the product of depth, width, and a measure of activation complexity (up to a scaling factor). This yields an identical role to the three sources of expressiveness: no exponential growth with depth is observed anymore.

 \end{abstract}


\section{Introduction}

The ability to train deep parametric models has enabled dramatic advances in a wide variety of fields, ranging from computer vision to natural-language processing \cite{bengio2009learning,krizhevsky2012imagenet}. Many popular deep models belong to the family of feedforward neural networks (NNs), for which the input-output mapping takes the form\footnote{\changer{In practice, the last layer of the NN is not necessarily an activation function. This case is also covered by our framework since $\V\sigma_L$ can be any continuous and piecewise-linear function, including the identity function.}}
\begin{equation}
\M x\mapsto (\V\sigma_L \circ \V f_{\V \theta_L} \circ  \V\sigma_{L-1} \circ \cdots \circ \V\sigma_2 \circ \V f_{\V \theta_2} \circ \V\sigma_1 \circ \V f_{\V \theta_1})(\M x),
  \end{equation}
  where $L$ is the number of layers of the NN (referred to as the depth of the NN), $\V f_{\V \theta_k}\colon\R^{d_k}\rightarrow\R^{d_{k+1}}$ is an affine function parameterized by $\V \theta_k$, and $\V \sigma_k$ is a non-affine activation function. One of the most widespread activation functions in deep learning is the rectified linear unit $\mathrm{ReLU}(x) = \max(x,0)$ \cite{glorotDeepSparseRectifier2011,maas2013rectifier,Lecun2015}. With this choice, the mapping is a composition of continuous and piecewise-linear (CPWL) functions, which yields a map that is CPWL too \cite{montufar2014number}. Remarkably, the reverse also holds true: any CPWL function $\mathbb{R}^d\rightarrow \mathbb{R}$ can be parameterized by a ReLU NN with at most $\lceil \log_2(d+1) \rceil$ hidden layers \cite{Arora2018}. The family of NNs generating CPWL functions (referred to as CPWL NNs in the sequel) is broad. It benefits from a large choice of effective activation functions that includes ReLU \cite{Lecun2015}, leaky ReLU \cite{maas2013rectifier}, PReLU \cite{He2015}, CReLU \cite{Shang2016}, Maxout \cite{Goodfellow2013}, linear splines \cite{agostinelliLearningActivationFunctions2015,unserRepresenterTheoremDeep2019}, GroupSort \cite{anil2019sorting}, Householder \cite{Singla} as well as other components such as convolutional layers, max- and average-pooling, skip connections \cite{heDeepResidualLearning2016}, and batch normalization \cite{ioffe2015batch}(once the model is trained). 
  While the depth of the architecture is instrumental to overcome the curse of dimensionality \cite{eldanPowerDepthFeedforward2016,mhaskarDeepVsShallow2016,poggioWhyWhenCan2017}, it concurrently deters our understanding of the parameterization when compared to simpler models \cite{Goujon2021}.\\

The observation that a ReLU NN produces a CPWL function sheds light on its behavior. In effect, a ReLU NN partitions the input domain into affine regions \cite{balestrieroSplineTheoryDeep2018,balestrieroMadMaxAffine2021}. The characteristics of the regions are therefore fundamental to grasp the structure of the learnt mapping and there exist different approaches to define them \cite{Tarela1990,Tarela1999}. The regions can be described as polyhedrons or union of polyhedrons, which results from the continuity and the piecewise-affine property of the mapping. In the case of ReLU NNs, it is common to define activation regions, which are sets of points that fire the same group of neurons. On each activation region, the mapping is affine and these sets are convex \cite{Hanin2019a}. Unfortunately, the linear regions in deep NNs are only indirectly specified. While they can be locally described \cite{Balesterio2019} their global delimitation becomes computationally less and less tractable as the dimension increases, which compromises the interpretability of deep NNs. Yet, it is entangled with their ability to overcome the curse of dimensionality.

The successive compositions inherent in deep models prevent us from attributing a specific role to each parameter. The size and the expressiveness of the function space $\mathcal{H}_{\mathcal{N}}$ generated by a given architecture $\mathcal{N}$ is consequently remotely connected to the number of trainable parameters. With their remarkable structure, CPWL NNs benefit from another meaningful descriptor: the distribution of counts of regions of all the mappings that the architecture can produce.
Two approaches have been proposed to give a better understanding of this descriptor.

\begin{itemize}
  \item {\em Upper and lower bound the maximum number of regions of the CPWL mappings generated by a given architecture.} \change{The first bounds, given in \cite{montufar2014number}, showed that the maximum number of regions that can be produced by ReLU NNs increases exponentially with their depth. This revealed that deep models have the ability to generate much more complex functions than shallow ones do. The bounds for ReLU NNs have since been refined, for example in \cite{Serra2018a} and then in \cite{hinz2019framework}, and also extended to other NN architectures. For instance, \cite{pmlr-v119-xiong20a} specifies bounds for the maximum number of regions of convolutional NNs (CNNs). It is shown that CNNs produce more regions per parameter than fully connected NNs do. For Maxout NNs, bounds can be derived directly from the ones on ReLU NNs \cite{montufar2014number, Serra2018a}. However, this approach usually yields loose bounds, as recently shown in \cite{MontuRenZhangMaxout2022}. The derivation of sharp bounds for Maxout NNs, as proposed in \cite{MontuRenZhangMaxout2022}, requires to take into account the specificities of the Maxout unit, and it was handled via the use of tropical geometry.}

  The available bounds show that the maximum number of regions in ReLU and Maxout NNs increases exponentially with their depth. It suggests that deep models have the ability to generate more complex functions than shallow ones do \cite{montufar2014number,Serra2018a,Arora2018,hinz2019framework, MontuRenZhangMaxout2022}.

 \item {\em Upper bound the average number of regions of the mappings generated by ReLU and Maxout NNs.} The available bound for ReLU NNs depends on the number of neurons, regardless of whether the NN is deep or wide, and depth does not produce exponentially more regions on average \cite{Hanin2019,Hanin2019a}. In other words, this behavior drastically differs from the maximum number of regions.\change{ This new perspective was then recently extended to Maxout NNs, with a similar qualitative conclusion \cite{TseranMontu2021expected}}.
\end{itemize}

\begin{figure}
  \centering
  \centerline{\includegraphics[width=120mm]{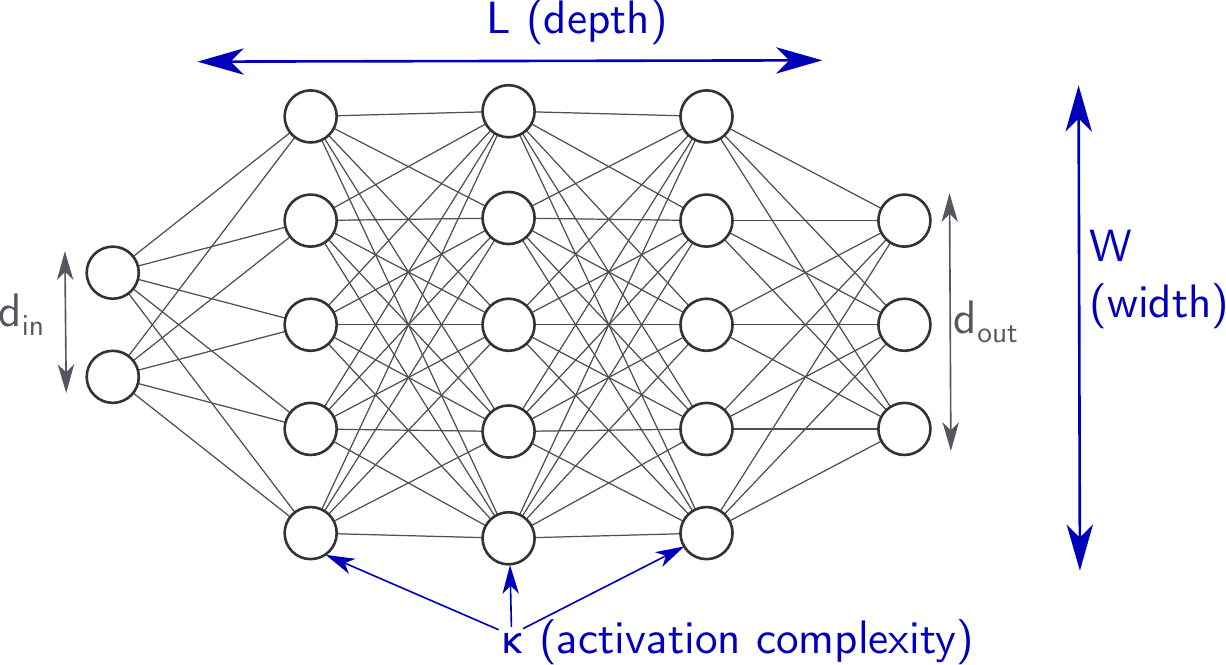}}
  \caption{The three sources of complexity of CPWL NNs.}
  \label{fig:CPWLNet_Descriptor}\medskip
\end{figure}

The existing toolbox of CPWL NNs is broad and likely not complete yet, as hinted by recent works on the MaxMin or more generally GroupSort activation function, in the field of Lipschitz-constrained NNs \cite{anil2019sorting,tanielian2021approximating}\change{, or with the Piecewise Linear Unit (PWLU) \cite{ZhuPWLU2023}.} Previous studies on the count of linear regions have provided insights on some specific CPWL NNs only, mostly ReLU and Maxout NNs. Their qualitative outcomes turn out to hold true for CPWL NNs in general. We intend to prove this claim with quantitative results in this paper. We want to improve the understanding of the role of the three main ways to increase the expressiveness of CPWL NNs (Figure \ref{fig:CPWLNet_Descriptor}), namely,
\begin{itemize}
  \item {\bf depth}, which is the number of composed CPWL functions;
  \item {\bf width}, which relates the input and output dimensions of the composed CPWL layers;
  \item {\bf activation complexity}, the rationale there being that the expressiveness of a CPWL NN can be heightened by increasing the complexity of its activation functions. This strategy is used with both univariate and multivariate activation functions, and it gave rise to deepspline, Maxout, GroupSort, and \change{PWLU} NNs for example. \change{In the remainder of the paper, the complexity of an activation will refer to its number of linear regions. For example, a rank-$k$ Maxout unit has a complexity of $k$ (see Figure \ref{fig:activationComplexity} for visual examples).}
\end{itemize}

Our contributions are as follows.
\begin{enumerate}[label=(\roman*)]
  \item Generalization of the notion of arrangement of hyperplanes to arrangement of convex partitions with analogous tight bounds on the number of regions.
  \item Determination of precise bounds on the maximal number of linear convex regions generated by the primary operations of the space of CPWL functions (sum, vectorization, and composition). The compositional upper and lower bound grow exponentially with depth and polynomially with the width and the activation complexity.
  \item Demonstration that, under reasonable assumptions, the expected number of regions along a 1D path for random CPWL NNs is at most linear with the product of the depth, the width, and the activation complexity (up to an independent factor), which yields equivalent roles to the three descriptors in terms of expressiveness.
\end{enumerate}
The paper is organized as follows: In Section \ref{sc:prelim}, we present the relevant mathematical concepts. In Section \ref{sc:deterministic}, we bound from below and from above the maximal number of regions produced by CPWL NNs and, in Section \ref{sc:stochastic}, we present a stochastic framework to quantify the average expressiveness of CPWL NNs with random parameters.
\section{Mathematical Preliminaries}
\label{sc:prelim}
\subsection{CPWL Functions}
\begin{figure}
  \centering
  \centerline{\includegraphics[width=150mm]{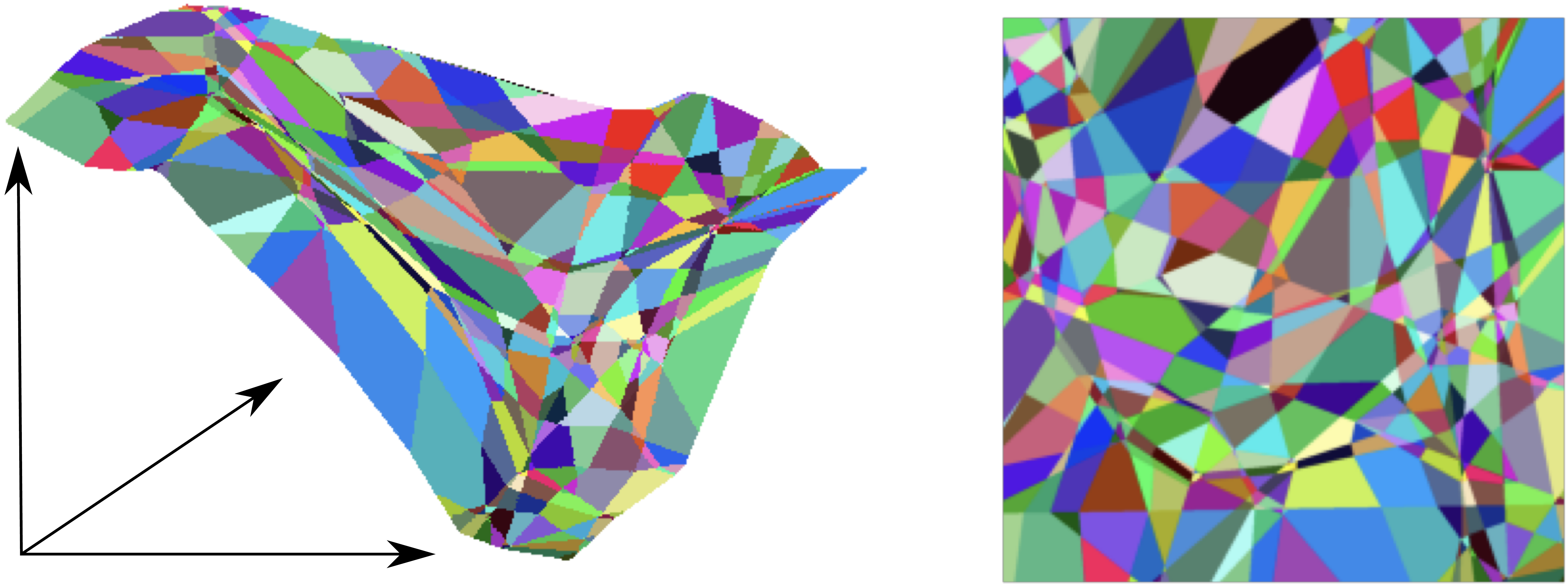}}
  \caption{An $\R^2\rightarrow\R$ CPWL function and its corresponding partition of the input space.}
  \label{fig:CPWL}\medskip
\end{figure}

\begin{definition}
  \label{df:cpwl}
  A function $\V f \colon \R^d \rightarrow \R^{d'}$ is continuous and piecewise-linear (CPWL) if it is continuous and if there exists a set $\{\V f^k\colon k\in\{1,\ldots,K\}\}$ of affine functions and closed subsets $(\Omega_k)_{k=1}^K$ of $\R^d$ with nonempty and pairwise disjoint interiors such that $\cup_{k=1}^K\Omega_k = \R^d$ and $\V f_{|\Omega_k}=\V f^k$ on $\Omega_k$. \change{The $\V f^k$ are called the affine pieces of $\V f$, and the $\Omega_k$ the corresponding projection regions.}
\end{definition}
An example of a CPWL function and of its partition is given in Figure \ref{fig:CPWL}. The $k$th component of a vector-valued CPWL function $\V f_\ell$, which is necessarily CPWL as well, will be denoted by $f_{\ell,k}$. The space of CPWL functions has the following remarkable properties:
  \begin{itemize}
    \item it is closed under compatible compositions;
    \item it is closed under compatible linear combinations;
    \item it is closed under compatible vectorization.
  \end{itemize}
 Since the function $\M x \mapsto \max(\M x)=\max(x_1,\ldots,x_d)$ is CPWL (with $d$ regions), the space of CPWL functions is also closed under max-pooling.
  \subsection{Regions of CPWL Functions and Convex Partitions}
\begin{figure}
  \centering
  \centerline{\includegraphics[width=150mm]{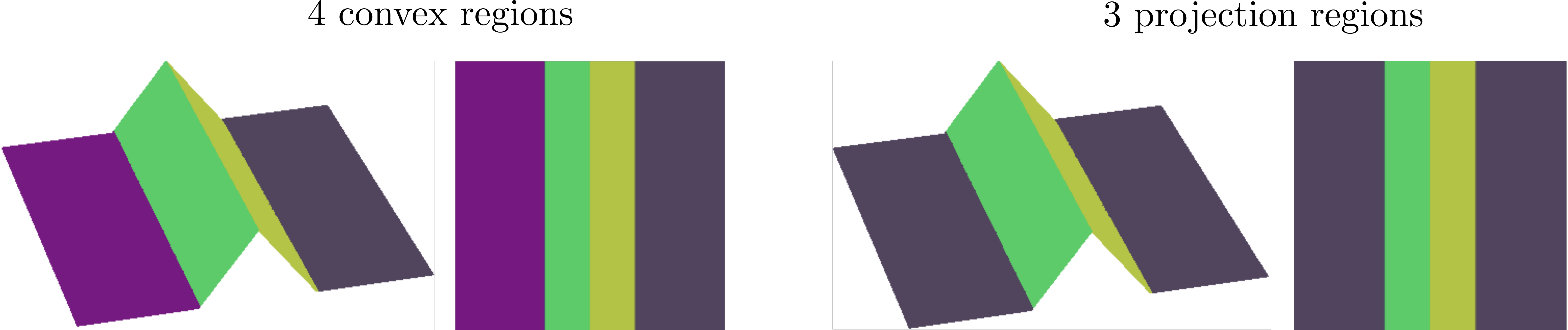}}
  \caption{Convex and projection regions of the CPWL function $(x,y)\mapsto \mathrm{ReLU}(\min(x+1,-x+1))$.}
  \label{fig:region}\medskip
\end{figure}

  \change{The term {\it linear region} is frequently used in an ambiguous way and may refer to different mathematical definitions. In the sequel, we shortly present some relevant definitions and discuss them in the context of CPWL NNs.}
  \change{
  \subsubsection{Projection Regions}
  }
  We recall that a polyhedron is the intersection of finitely many half-spaces, and that a polytope is a bounded polyhedron. The subsets $\Omega_k$ in Definition \ref{df:cpwl} are commonly referred to as projection regions \cite{Tarela1990,Tarela1999}. The affine pieces of different projection regions are distinct and, since the overall function is continuous, the common points of two neighboring regions lie in a hyperplane. This implies that the $\Omega_k$ are polyhedrons or unions of polyhedrons. These projection regions might, however, not be connected (Figure \ref{fig:region}).
    \change{
  \subsubsection{Convex Regions}
  }
 \change{It is usually preferred to work with (connected) convex regions because of their simpler geometrical structure. We now precisely define convex linear regions of CPWL functions.}
   \begin{definition}[Convex partitions of $\mathbb{R}^d$, adapted from \cite{Leon2018}]
    \label{df:convexparition}
    Let $n$ and $d$ be two positive integers. A convex partition of $\;\mathbb{R}^d$ is a collection $\partition = (P_1, P_2, . . . , P_n)$ of convex and closed subsets of $\mathbb{R}^d$ with nonempty and pairwise-disjoint interiors so that the union $\bigcup_{k=1}^n P_k = \mathbb{R}^d$. Each of the sets $P_k$ is called a region of $\partition$. Convex partitions with $n$ regions are called $n$-partitions.
    \end{definition}
    \change{
    \begin{definition}[Linear convex partition]
        A convex partition $\partition$ of $\R^d$ is said to be a linear convex partition of a CPWL function $\V f \colon\R^d\rightarrow \R^{d'}$ if $\V f$ is affine on each region of $\partition$.
    \end{definition}}
    The existence of a linear convex partition is guaranteed for any CPWL function but not its unicity. \change{This motivates Definition \ref{ref:nbcvxregions}, which gives a precise meaning to the {\it number of convex linear regions} for CPWL functions.
    \begin{definition}[Number of convex linear regions]
    \label{ref:nbcvxregions}
        The number $\reg_{\V f}$ of convex linear regions of $\V f$ is defined as the minimal cardinality of all linear convex partitions of $\V f$.
    \end{definition}}
    
A special instance of the linear convex regions for scalar-valued CPWL functions are the uniquely-ordered regions. Each of these regions has the same ordering of the values of the affine pieces $f^k$ of $f$ in all its points \cite{Tarela1999}. Uniquely-ordered regions are used to build the lattice representation of a CPWL function \cite{Tarela1990} and are tightly connected to the GroupSort activation function \cite{anil2019sorting}.

\subsubsection{Projection vs Convex Linear Regions}
\change{In the remainder of the paper we shall keep in mind the following connections between projection and convex linear regions.}
 \begin{itemize}
     \item Projection regions can always be partitioned into convex regions so that any upper bound on the number of convex regions also applies to the number of projection regions. Conversely, the number of convex regions can also be upper bounded by the number of projection regions (Proposition \ref{pr:projectionvsconvex}).
     \item 
     The majority of commonly used parameterizations have typically convex projection regions. The local parameterization with hat basis functions produces simplicial linear splines whose natural regions are simplices \cite{Goujon2021} and, therefore, are convex. Other known linear expansions, such as the generalized hinging-hyperplanes model \cite{Wang2005}, use nonlocal CPWL basis functions that partition the input domain into convex regions. The generated function will produce projection regions that are convex for all sets of parameters except for some specific values that are usually encountered with zero probability in a learning framework. The convex regions are also naturally adapted to compositional models such as ReLU and Maxout NNs as explained it \cite{Hanin2019}. 
 \end{itemize}
 
\begin{proposition}
  \label{pr:projectionvsconvex}
  Let $\V f\colon\R^d\rightarrow\R^{d'}$ be a CPWL function with $\rho$ projection regions. The number $\reg$ of linear convex regions of $\V f$ is no larger than the number of convex regions formed by the arrangement of $\rho(\rho-1)/2$ hyperplanes
  \begin{equation}
    \rho\leq \reg \leq \begin{cases}
      2^{\rho(\rho-1)/2},& \rho(\rho-1)/2 \leq d\\
      \sum_{k=0}^{d} {\rho(\rho-1)/2 \choose k},&\text{otherwise}.
    \end{cases}
  \end{equation}
\end{proposition}
\change{The proof of Proposition \ref{pr:projectionvsconvex} is given in \ref{ap:proofsec2}.}

\subsubsection{Useful Properties of Convex Partitions}
 \change{We now give a series of lemmas on convex partitions that are used in the proofs of Section \ref{sc:deterministic}. The proofs are given in \ref{ap:proofsec2}.}

 For convenience, we extend the definition of convex partitions of $\R^d$ to convex partitions of affine subspaces of $\R^d$. In particular, a convex partition of an affine subspace $E$ of $\R^d$ consists of convex and closed subsets of $E$ of dimension $\dim(E)$ whose pairwise intersection is of dimension smaller than $\dim(E)$ and whose union is $E$.
 
\begin{lemma}[Projection of a convex partition]
  \label{lm:subpartition}
  Let $E$ be an affine subspace of $\R^d$ and $\partition$ an $n$-partition of $\mathbb{R}^d$. Then, there exists a convex partition $\partition_E$ of $E$ in $\R^{d}$ with no more than $n$ regions such that, for $P_E\in \partition_E$, there is $P\in \partition$ with $P_E \subset P$.
  \end{lemma}

\begin{lemma}[Preimage of a convex partition under affine maps]
  \label{lm:inverseimagepartition}
  Let $\V f \colon \mathbb{R}^d \rightarrow \mathbb{R}^{d'}$ be an affine function and $\partition$ be an $N$-partition of the affine space $\V f(\mathbb{R}^d)$ in $\R^{d'}$. Then, $\V f^{-1}(\partition) = \{\V f^{-1}(P) \colon P\in\partition\}$ is an $N$-partition of $\mathbb{R}^d$.
  \end{lemma}

  \begin{lemma}
    \label{lm:dim}
    Let $(\V f_\ell)_{\ell\in [L]}$ be a collection of affine functions with $\V f_\ell:\mathbb{R}^{d_{\ell}}\rightarrow\mathbb{R}^{d_{\ell+1}}$ .Then,
    \begin{equation}
     \mathrm{dim}((\V f_L\circ \cdots \circ \V f_1)(\mathbb{R}^{d_1}))\leq \min{(d_1,\ldots,d_{L+1})}.
    \end{equation}
    \end{lemma}

\subsection{Arrangement of Convex Partitions}
\begin{figure}
    \centering
    \centerline{\includegraphics[width=150mm]{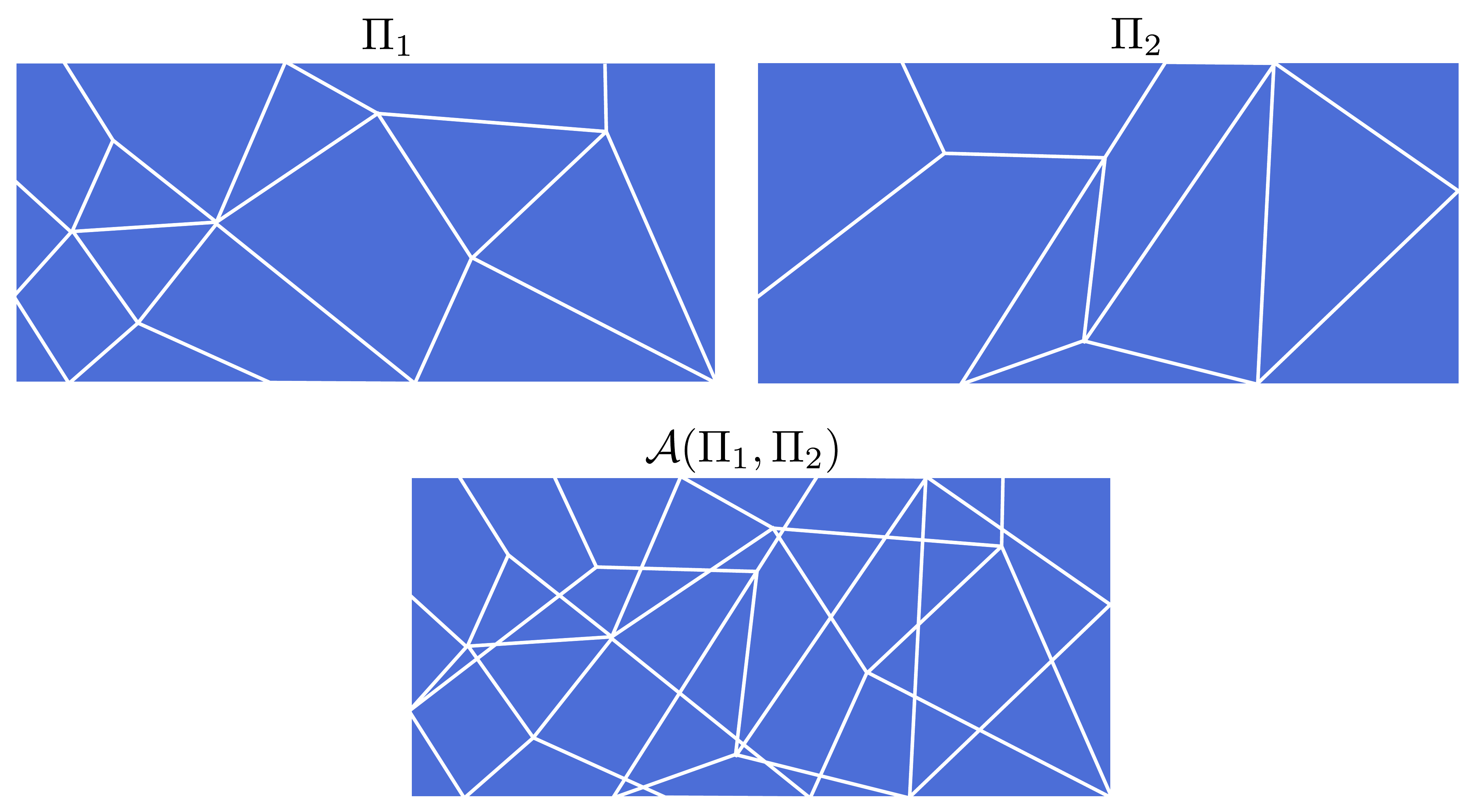}}
    \caption{Arrangement of two convex partitions of $\R^2$.}
    \label{fig:arrangement}\medskip
\end{figure}

The known results on the number of convex regions of ReLU NNs are built upon the theory of hyperplane arrangements. In combinatorial geometry, an arrangement of hyperplanes refers to a set of hyperplanes. It is known that the number of connected regions formed by an arrangement of $N$ hyperplanes in $\R^d$ is at most $\sum_{k=0}^{\min(d,N)}{N\choose k}$ \cite{zaslavsky1975facing}. This bound is reached when the hyperplanes are in general position: any collection of $k$ of them intersect in a $(d-k)$-dimensional plane for $1\leq k\leq d$ and have an empty intersection for $k>d$. Although this positioning seems very specific, it is qualified as ``general'' because it almost surely happens when the hyperplanes are randomly generated (with a ``reasonable'' notion of randomness).
When it comes to the study of generic CPWL NNs, the concept of arrangement of hyperplanes lacks precision since only a small fraction of all convex partitions can be seen as arrangement of hyperplanes. We thus introduce the notion of arrangement of convex partitions (Definition \ref{df:arrangement} and Figure \ref{fig:arrangement}) as a generalization, which will prove to be necessary to find the precise bounds given in Section \ref{sc:deterministic}. Note that, in the case of an arrangement of $N$ hyperplanes, our terminology differs. Instead of considering the hyperplanes, we rather consider the $N$ 2-partitions they form, which consist of pairs of closed half-spaces separated by the hyperplanes.

    \begin{definition}[Arrangement of convex partitions]
  \label{df:arrangement}
  Let $(\partition_k)_{k\in [N]}$ be a collection of $N$ convex partitions, with $\partition_k = (P^k_1,\ldots,P^k_{n_k})$, for $k\in[N]$. The arrangement $\mathcal{A}(\partition_1,\ldots,\partition_N)$ of these partitions is the convex partition whose regions are the \change{$A_{m_1, \ldots, m_N}$ that have nonempty interiors, where
  $$ A_{m_1, \ldots, m_N}=\bigcap_{k=1}^N P_{m_k}^k,$$  for $(m_1, \ldots, m_N) \in \{ 1, \ldots, n_1\}\times \cdots\times \{ 1, \ldots, n_N\}$.}
  \end{definition}

\section{Maximum Number of Regions Produced by CPWL NNs}
\label{sc:deterministic}
In this section, we characterize the largest number of regions that can be generated by simple operations with CPWL functions, including sums, vectorizations, and compositions. \change{In particular, we strictly generalize the known upper and lower bounds on the number of regions of ReLU NNs \cite{Serra2018} and Maxout NNs \cite{MontuRenZhangMaxout2022} to NNs activated by generic CPWL activation functions.}
\subsection{Upper Bound on the Number of Regions of Arrangements}
\begin{figure}
  \centering
  \centerline{\includegraphics[width=120mm]{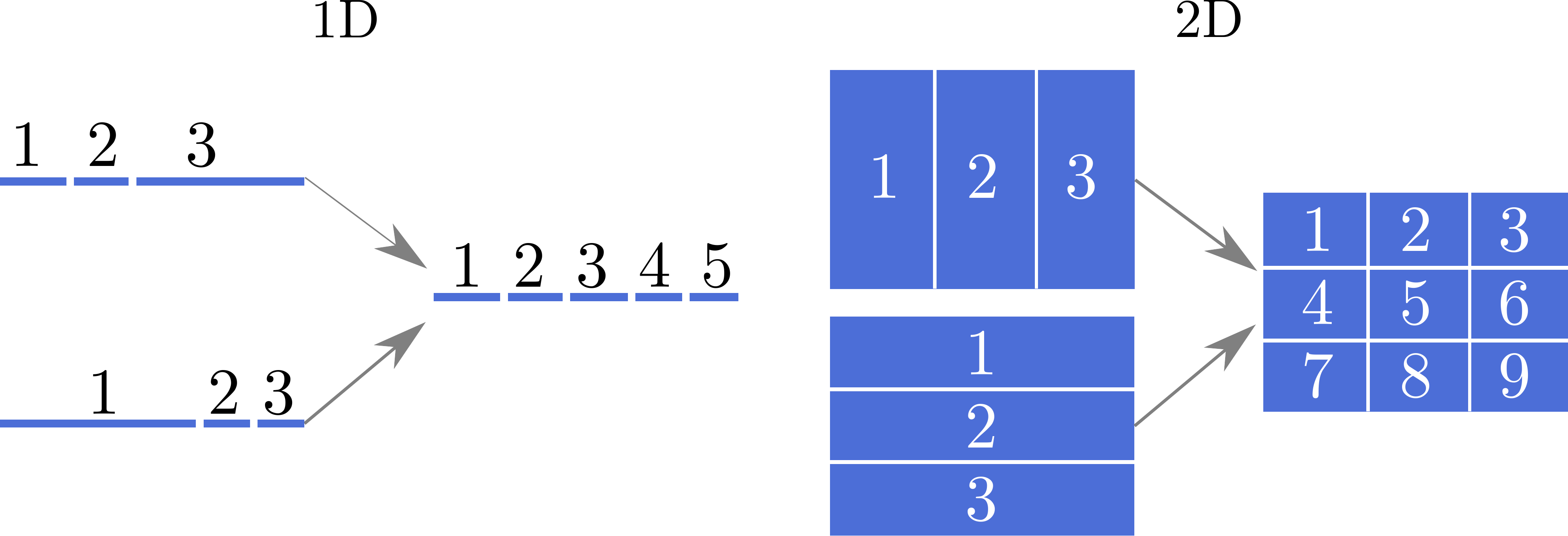}}
  \caption{Arrangement of two convex partitions with $3$ regions each. While in 2D the maximal number of regions is $3\times 3 = 9$, this cannot be reached in 1D, for which the maximum is $5$.}
  \label{fig:arrangementstaturation}\medskip
\end{figure}

Operations with CPWL functions imply arrangements of convex partitions, either explicitly, for sums and vectorizations, or implicitly, for compositions. It is straightforward to see that an arrangement $\mathcal{A}(\partition_1,\ldots,\partition_N)$ of $N$ convex partitions $\partition_1,\ldots,\partition_N$ of $\R^d$ with $n_1,\ldots,n_N$ regions cannot yield more than $n_1 n_2 \cdots n_N$ regions. This naive bound is a polynomial of degree $N$ in $n_1,\ldots,n_N$. In dimension $d=1$ one can, however, check that the bound is not sharp: the number of regions is no more than $1+(n_1-1)+\cdots+(n_N-1)$. More generally, the number of regions of the arrangement is bounded by a polynomial in the cardinality of the partitions $n_1,\ldots,n_N$ of degree $\min(d,N)$ (Theorem \ref{th:upperbound}), which highlights the role played by the dimension of the ambient space.
\begin{theorem}[Arrangements' upper bound]
  \label{th:upperbound}
  The maximum cardinality $\ub^d(n_1,\ldots,n_N)$ of the arrangement $\mathcal{A}(\partition_1,\ldots,\partition_N)$ of $N$ convex partitions $\partition_1,\ldots,\partition_N$ of $\R^d$ with cardinality $n_1,\ldots,n_N$ is a polynomial in $n_1,\ldots,n_N$ of degree $\min(d,N)$. It is given by
\begin{equation}
  \ub^d(n_1,\ldots,n_N) = 1 + \sum_{k=1}^{\min(d,N)} \sum_{1\leq \ell_1<\cdots<\ell_k\leq N} \prod_{q=1}^k(n_{\ell_q}-1).
\end{equation}
   Moreover, this bound satisfies
    \begin{equation}
      \begin{matrix*}[l]
      \ub^d(n_1,\ldots,n_N) &= \prod_{k=1}^N n_k, &\text{ if } N\leq d\\
    \ub^d(n_1,\ldots,n_N) &\leq \left(1+\sum_{k=1}^N (n_k-1)\right)^d \leq \left(\sum_{k=1}^N n_k\right)^d, &\text{ otherwise.}
  \end{matrix*} \end{equation}   
\end{theorem}
 The expression of the bound in Theorem \ref{th:upperbound} is based on a broad result of discrete geometry \cite{Bulavka2020}. We then relied on Zaslavsky's Theorem \cite{zaslavsky1975facing} and Whitney's formula to construct a specific arrangement of convex partitions for any set of parameters $d,N,n_1,\ldots,n_N \inN\backslash\{0\}$ that achieves the bound. The proof of Theorem \ref{th:upperbound} is given in \ref{ap:proofsec2}. We now discuss the result and its implications.
 \begin{itemize}
  \item Theorem \ref{th:upperbound} is a generalization of the hyperplane-arrangement bound. Indeed, let us consider the number of regions generated by an arrangement of $N$ hyperplanes: each hyperplane defines a 2-partition of $\R^d$ and the bound yields $\ub^d(2,\ldots,2) = 1 + \sum_{k=1}^{\min(d,N)} \sum_{1\leq \ell_1<\cdots<\ell_k\leq N} 1=1 + \sum_{k=1}^{\min(d,N)} {N \choose k}=\sum_{k=0}^{\min(d,N)} {N \choose k}$, which is known to be exactly the number of convex regions generated by an arrangement of $N$ hyperplanes in general position \cite{zaslavsky1975facing}.
   \item The naive upper bound can be rewritten as \\$\prod_{k=1}^N n_k = \prod_{k=1}^N ((n_k-1)+1) =  1 + \sum_{k=1}^{N} \sum_{1\leq \ell_1<\cdots<\ell_k\leq N} (n_{\ell_1}-1) \cdots (n_{\ell_k}-1)$. This shows that when $N\leq d$, the naive bound is optimal. By contrast, when $N>d$, the dimension enforces the existence of one or more empty intersections between regions of different partitions. This is illustrated in Figure \ref{fig:arrangementstaturation} with a simple example.
   \item For partitions with the same number $n$ of regions, we introduce the simpler notation $\ub^d_N(n) \coloneqq \ub^d(n,\ldots,n)= 1 + \sum_{k=1}^{\min(d,N)} \sum_{1\leq \ell_1<\cdots<\ell_k\leq N} (n-1)^{k} = \sum_{k=1}^{\min(d,N)} {N\choose k}(n-1)^{k}\leq \min(n^N,(1+N(n-1))^d)$.
   \item The bound is reached when the partitions $\partition_k$ are made of the regions of the arrangement of $(n_k-1)$ distinct parallel hyperplanes, where the hyperplanes are in general position when only one per partition is selected (more detailed in the proof in \ref{ap:proofsec2}).
 \end{itemize}
 
\begin{remark}
After the disclosure of our work on arXiv, we became aware of \cite{MontuRenZhangMaxout2022}, which contains highly relevant results on the complexity of Maxout NNs. From Theorem \ref{th:upperbound}, we can directly recover the sharp bound on the number of regions of a shallow Maxout NN recently given in \cite[Theorem 3.7]{MontuRenZhangMaxout2022}. Regarding the converse, i.e. inferring Theorem \ref{th:upperbound} from \cite[Theorem 3.7]{MontuRenZhangMaxout2022}, we believe that it could perhaps be done but it is not immediate. Indeed, \cite[Theorem 3.7]{MontuRenZhangMaxout2022} is specific to convex partitions that are the linear partitions of the maximum of affine functions, i.e. only specific CPWL functions. Note that the proofs in \cite{MontuRenZhangMaxout2022} rely on tropical geometry, which gives an overall perspective very different from ours.
\end{remark}

\subsection{Single Hidden-Layer: Bound for the Sum and Vectorization Operations}
The sum and vectorization of CPWL functions both yield the same bound on the number of linear convex regions. \change{We now give a novel optimal bound in Proposition \ref{pr:sumandvectorization}. The proof is given in \ref{ap:proofsec2}.}
\begin{proposition}
  \label{pr:sumandvectorization}
  Let $f_1,\ldots,f_{N}\colon \R^d\rightarrow\R$ be CPWL functions with $\reg_1,\ldots,\reg_N$ convex linear regions. The number of convex linear regions of the sum $(f_1+\cdots+f_N)$ and of the vector-valued function $(f_1,\ldots,f_N)$ can be bounded by a polynomial in $\reg_1,\ldots,\reg_N$ of degree $\min(d,N)$, namely
  \begin{align}
    \reg_{f_1+\cdots+f_N}&\leq \ub^d(\reg_1,\ldots,\reg_N),\\
    \reg_{(f_1,\ldots,f_N)}&\leq \ub^d(\reg_1,\ldots,\reg_N),
  \end{align}
  and these bounds are sharp.
\end{proposition}

\begin{remark}
    Bounds similar to the ones given in Proposition \ref{pr:sumandvectorization} have recently been derived for one hidden-layer Maxout NNs \cite{MontuRenZhangMaxout2022}. The latter work is a specific instance of our setting, in which the CPWL functions considered are the maximum of a finite set of affine functions.
\end{remark}
As an illustration of Proposition \ref{pr:sumandvectorization} and Theorem \ref{th:upperbound}, we give some direct implications on the number of regions of some building blocks of CPWL NNs before going deeper.
\paragraph{Ridge Functions} Consider the ridge expansion $f_R\colon \M x\mapsto \sum_{k=1}^N \lambda_k \mathrm{ReLU}(\M w_k^T \M x + b_k)$, where $\M w_k\in\R^d$ and $b_k\inR$. The number $\reg_{\mathrm{Ridge}}$ of linear convex regions of $f_R$ is upper-bounded as
\begin{equation}
  \reg_{\mathrm{Ridge}}\leq \ub^{d}_N(2) = \sum_{k=0}^{\min(d,N)} {N\choose k}\leq \min(2^N,(N+1)^d),
\end{equation}
and the bound is tight.
\paragraph{Max-Pooling} The $k$th component of the max-pooling operation $\V f_\mathrm{mp}\colon\R^d\rightarrow\R^{d'}$ takes the form $\V f_\mathrm{mp}^k(x_1,\ldots,x_d) = \max_{p\in I_k}(x_p)$, where $I_k$ is a set of chosen cardinality $N$ of ``neighboring'' coordinate indices. The number $\reg_{\mathrm{mp}}$ of convex linear regions of the max-pooling operation is upper-bounded as
\begin{equation}
  \reg_{\mathrm{mp}}\leq \ub^{d}_{d'}(N) = \sum_{k=0}^{\min(d,d')} {d'\choose k}(N-1)^k.
\end{equation}

\paragraph{Generalized Hinging Hyperplanes (GHH)}Consider the GHH expansion $\R^d\rightarrow\R$ $f_G = \sum_{k=1}^N \epsilon_k \max(f_1^k,\ldots,f_{d+1}^k)$, where $f_{p}^k$ are affine functions and $\epsilon_k=\pm 1$ \cite{Wang2005}. The number $\reg_{\mathrm{GHH}}$ of convex linear regions of $f_G$ is upper-bounded as
\begin{equation}
  \reg_{\mathrm{GHH}}\leq \ub^{d}_N(d+1) = \sum_{k=0}^{\min(d,N)} {N\choose k}d^{k}\leq \min\left((d+1)^N,(Nd+1)^d\right).
\end{equation}

\paragraph{GroupSort Layer} The sort operation takes as input a vector $\M x\inR^d$ and simply sorts its components. For any permutation $\sigma$ of the set $\{1,\ldots,d\}$, we define the uniquely-ordered region $P_{\sigma}=\{\M x\in\R^d\colon x_{\sigma(1)}\leq \cdots\leq x_{\sigma(d)}\}$, where $x_k$ is the $k$th component of $\M x$. These regions are convex as intersections of half-spaces and the sort operation agrees on them with distinct affine functions, namely, permutations. We infer that the sort operation has exactly $d!$ linear convex regions and the same number of projection regions.

The GroupSort activation was recently introduced and shown to be beneficial in the context of Lipschitz-constrained learning \cite{anil2019sorting}. It generalizes the minmax and sort activations: it splits the pre-activation into a chosen number $n_g$ of groups of size $g_s$ (with $n_g g_s=d$), sorts each pre-activation of each group in ascending order, and outputs the combined sorted groups. Each group produces $g_s!$ linear convex regions which are invariant along the coordinates that are not in the group. We infer the number of linear convex regions of the GroupSort activation to be $\reg_{\mathrm{GS}} = (g_s !)^{n_g}$, which can be bounded as
\begin{equation}
  (g_s/2)^{d/2}\leq \reg_{\mathrm{GS}}\leq (g_s^{g_s})^{n_g} = g_s^d,
\end{equation}
where we have used the known inequalities $(n/2)^{n/2}\leq n ! \leq n^n$. The bounds support the intuition that larger group sizes generate more regions than smaller ones. Note, however, that they simultaneously increase the computational complexity of the layer.

\change{
\paragraph{PWLU}
The PWLU \cite{ZhuPWLU2023} is a learnable CPWL activation function with control points placed on a grid and with fixed linear regions (namely simplices whose vertices are control points). In its 2D version, a PWLU $ \varphi_{\mathrm{PWLU}}\colon \R^2 \rightarrow \R$ with $M^2$ control points has $2(M-1)^2$ linear regions that are triangles, see Figure \ref{fig:activationComplexity} for an illustration with $M=4$, and see \cite[Figure 5]{ZhuPWLU2023} for a more generic representation of PWLUs. Consider the one-hidden layer $\R^d\rightarrow \R$ PWLU NN $f_{\mathrm{PWLU}}(\M x)= \sum_{k=1}^{N}\varphi^k_{\mathrm{PWLU}}(\M W_k \M x)$ with 2D PWLU activations $\varphi^k_{\mathrm{PWLU}}$ with $M^2$ control points and corresponding weight matrices $\M W_k \inR^{2\times d}$. The number $\kappa_{\mathrm{PWLU}}$ of convex linear regions of this PWLU NN is upper-bounded as
\begin{align}
    \kappa_{\mathrm{PWLU}} \leq \beta^d_N\left(2(M-1)^2\right) &= \sum_{k=1}^{\min(d,N)} {N\choose k}(2(M-1)^2-1)^{k}\\
    &\leq \min((2(M-1)^2)^{N},(1+N(2(M-1)^2-1))^d)\\
    &\leq \min((2M^2)^{N},(1+N(2M^2))^d).
\end{align}
Our framework also allows one to derive bounds for NNs activated with higher dimensional PWLUs, but we are not aware of their use in practice.
}
\subsection{Multiple Hidden-Layers: Compositional Bounds}
\begin{figure}
  \centering
  \centerline{\includegraphics[width=\linewidth]{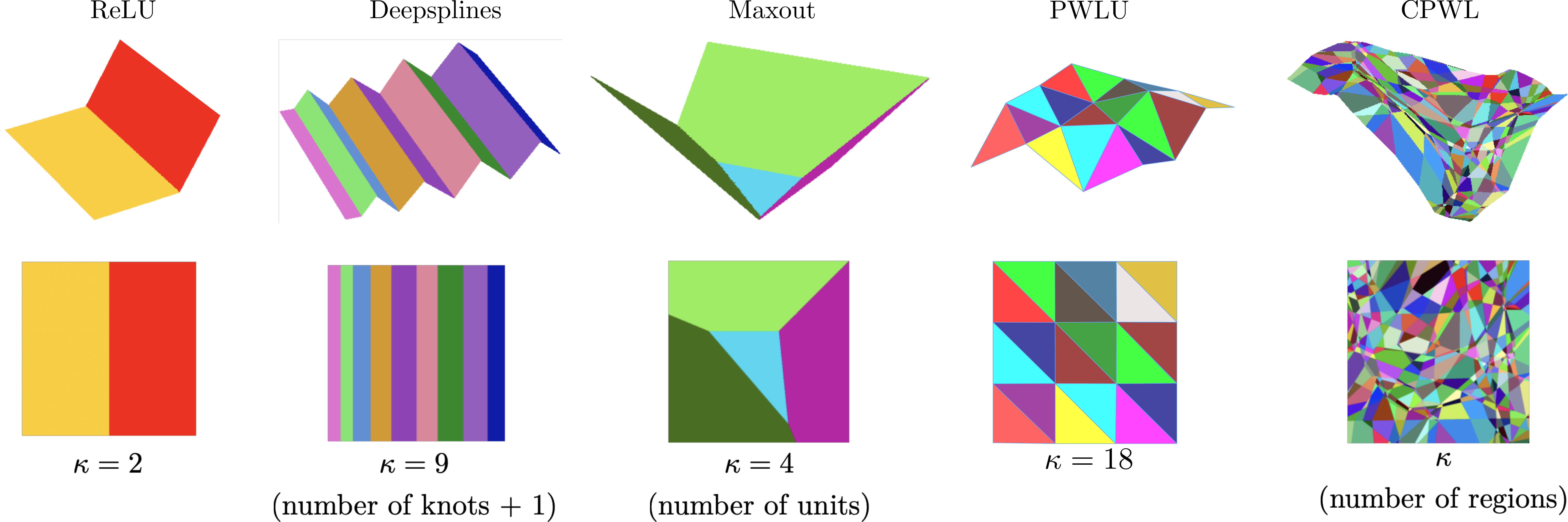}}
  \caption{Partition and complexity of some CPWL components.}
  \label{fig:activationComplexity}\medskip
\end{figure}

The architecture of a CPWL NN $\R^{d_1}\rightarrow\R^{d_{L+1}}$ is specified by its depth $L$, its layer dimensions $(d_1,\ldots,d_{L+1})$, and its activation complexity $\kappa_{\ell,k}$ at each node $(\ell,k)$, which is naturally depicted by the number of linear convex regions of the $k$th component of the $\ell$th composed function (Figure \ref{fig:activationComplexity}).
Theorem \ref{th:boundcomposition} below yields precise bounds on the maximal number of convex linear regions of any CPWL NN. It is complemented by Corollary \ref{cr:boundcompo} which tackles the following question: given a CPWL NN with fixed input and output dimensions, how is the maximal number of regions related to depth, width, and activation complexities? Our results confirm and generalize the following qualitative intuitions:
\begin{itemize}
  \item (i) depth can exponentially increase the complexity of the generated function;
  \item (ii) width and activation complexity, on the contrary, can only increase the number of linear convex regions of the generated function polynomially;
  \item (iii) layers with small dimensions reduce the maximal number of regions produced by the NN, especially if they are located toward the input of the NN. This bottleneck effect stems from the upper bound given in Theorem \ref{th:upperbound}.
\end{itemize}
Note that (i) is well known and was first proven in \cite{montufar2014number}, (ii) is in agreement with the recent results in \cite{MontuRenZhangMaxout2022} obtained for the particular instance of Maxout NNs, and (iii) was observed for ReLU NNs in \cite{MontufarNote2017, Serra2018a}.

\begin{theorem}
  \label{th:boundcomposition}
  The maximal number $\reg_{\max}$ of convex linear regions of a CPWL NN with depth $L$, layer dimensions $(d_1,\ldots,d_{L+1})$, and activation complexities $\kappa_{\ell,k}$ for $k=1,\ldots,d_{\ell+1}$ and $\ell=1,\ldots,L$, is bounded as
  \begin{equation}
    \prod_{\ell=1}^L \alpha^{\min(d_1,\ldots,d_{L+1})}(\reg_{\ell,1},\ldots,\reg_{\ell,{d_{\ell+1}}}) \leq \reg_{\max} \leq \prod_{\ell=1}^L \ub^{\min(d_1,\ldots,d_{\ell})}(\reg_{\ell,1},\ldots,\reg_{\ell,{d_{\ell+1}}}),
  \end{equation}
  where $\ub^{\cdot}(\cdot)$ is the upper bound on the number of regions of an arrangement of convex partitions (Theorem \ref{th:upperbound}) and where
  \begin{equation}
    \alpha^{\min(d_1,\ldots,d_{L+1})}(\reg_{\ell,1},\ldots,\reg_{\ell,{d_{\ell+1}}}) = \max_{\tau \in \mathcal{T}_{d_{\ell}}}\prod_{r=1}^{\min(d_1,\ldots,d_{L+1})}\sum_{k\in\tau^{-1}(\{r\})}\reg_{\ell,k}.
  \end{equation}
  There, $\mathcal{T}_{d_{\ell}}$ denotes the set of \change{all mappings from $\{k\inN\colon 1\leq k \leq d_{\ell+1}\}$ to $\{k\inN\colon 1\leq k \leq \min(d_1,\ldots,d_{L+1})\}$, and $\tau^{-1}(\{r\})$ denotes the preimage of $\{r\}$ under $\tau$}.
\end{theorem}

\begin{corollary}
  \label{cr:boundcompo}
  The maximal number $\kappa_{\max}$ of convex linear regions of a CPWL NN with $L$ layers, layer dimensions $(d_{\mathrm{in}},W,\ldots,W,d_{\mathrm{out}})$, with $d_{\mathrm{in}},W,d_{\mathrm{out}}\inN\backslash\{0\}$ and $W\geq d_{\mathrm{in}}$, where each component of the composed functions has $\kappa$ linear convex regions is bounded as
  \begin{equation}
    (\kappa \floor{W/d^*})^{Ld^*}\leq \kappa_{\max} \leq (\kappa W)^{Ld_{\mathrm{in}}},
  \end{equation}
  where $d^*=\min(d_{\mathrm{in}},d_{\mathrm{out}})$.
\end{corollary}
\begin{corollary}
  \label{cr:boundcompoprojection}
  The bounds given in Theorem \ref{th:boundcomposition} and Corollary \ref{cr:boundcompo} also apply to the maximal number of projection regions of a CPWL NN and, equivalently, to its maximal number of distinct affine pieces.
\end{corollary}
The proofs of Theorem \ref{th:boundcomposition} and of its corollaries can be found in \ref{ap:compositionalbounds}.

\subsection{Application to Some Popular CPWL NNs}
In the sequel, we consider the CPWL NN $\V f_L\circ\cdots\circ \V f_1$  where $\V f_\ell\colon\R^{d_\ell}\rightarrow\R^{d_{\ell+1}}$. We now apply Theorem \ref{th:boundcomposition} to bound the maximal number of convex linear regions produced by the most popular architectures. Note that the lower bound given in Theorem \ref{th:boundcomposition} only applies to CPWL NNs with pointwise activation functions. This includes ReLU and, more generally, deepspline NNs. The reason is that the lower bound of Theorem \ref{th:boundcomposition} was found by building a deepspline NN.
\paragraph{ReLU/PReLU/Leaky ReLU NNs}
In a ReLU NN, the $k$th component $f_{\ell,k}$ of $\V f_\ell$ takes the form $f_{\ell,k} \colon\M x\mapsto \mathrm{ReLU}(\M w_{\ell,k} \M x + b_{\ell,k})$ and has two convex linear regions (half-spaces). Theorem \ref{th:boundcomposition} then yields
\begin{equation}
  \label{eq:ReLUBoundDeter}
  \reg_{\rm ReLU}\leq \prod_{\ell=1}^L \sum_{k=0}^{\min(d_1,\ldots,d_\ell)}{d_{\ell+1}\choose k},
\end{equation}
which is the bound proposed in \cite{montufar2014number}. However, it is not the tightest upper bound known \cite{Serra2018a}. The reason is that the ReLU function is only a very specific instance of 1D CPWL functions with 2 linear regions: the image of the half real line $(-\infty,0]$ by the $\mathrm{ReLU}$ function is only the singleton $\{0\}$. This reduces the apparent dimension of the problem for any region that would not fire all neurons. This observation was exploited in \cite{Serra2018a} to get a better estimate. In that sense, \eqref{eq:ReLUBoundDeter} is better tailored to PReLU and Leaky ReLU NNs, which have activations with two nonzero slope regions.
\paragraph{Deepspline NN}
Deepspline NNs have learnable pointwise 1D CPWL activation functions \cite{agostinelliLearningActivationFunctions2015,unserRepresenterTheoremDeep2019,bohraLearningActivationFunctions2020}. Given activation functions with $(\kappa-1)$ knots (at most $\kappa$ linear convex regions), the number of linear convex regions of the NN is bounded as
\begin{equation}
  \label{eq:DSBoundDeter}
  \reg_{\rm Deepspline}\leq \prod_{\ell=1}^L \sum_{k=0}^{\min(d_1,\ldots,d_\ell)}{d_{\ell+1}\choose k}(\kappa-1)^k.
\end{equation}
\paragraph{Maxout NN}
In a Maxout NN with $\kappa$ units, the $k$th component $f_\ell^k$ of $\V f_\ell$ takes the form $f_\ell^k \colon\M x\mapsto \max(h_{\ell,k}^1,\ldots,h_{\ell,k}^\kappa)$, where $h_{\ell,k}^1,\ldots,h_{\ell,k}^\kappa$ are learnable affine functions \cite{Goodfellow2013}. Theorem \ref{th:boundcomposition} yields that
\begin{equation}
  \label{eq:MaxoutBoundDeter}
  \reg_{\rm Maxout}\leq \prod_{\ell=1}^L \sum_{k=0}^{\min(d_1,\ldots,d_\ell)}{d_{\ell+1}\choose k}(\kappa-1)^k.
\end{equation}
This bound is an improvement over \cite{Serra2018a}. In their work they plug $d_\ell=d$ for $\ell=1,\ldots,L$ and obtain the bound $2^{\frac{\kappa(\kappa-1)}{2}dL}$, to be compared to $\kappa^{dL}$ for \eqref{eq:MaxoutBoundDeter}.
\paragraph{GroupSort NNs}
To bound the number of linear convex regions of a GroupSort NN \cite{anil2019sorting} with the same group size $g_s$ in each layer, we consider for each composition the arrangement of $d_{\ell+1}/g_s$ convex partitions (one per group) with $g_s!$ regions each and obtain that
\begin{equation}
  \label{eq:GroupSortBoundDeter}
  \reg_{\rm GroupSort}\leq \prod_{\ell=1}^L \sum_{k=0}^{\min(d_1,\ldots,d_\ell)}{d/g_s\choose k}(g_s! - 1)^k.
\end{equation}
These bounds provide an intuition of the role of the hyperparameters of CPWL NNs in terms of expressiveness. For instance, the number of units in Maxout NNs plays a role in the bound that is analogous to that of the number of knots of the activation functions in deepspline NNs. However, these two architectures do not induce the same implementation complexity. To increase the activation complexity by one unit, Maxout requires the inclusion of an additional learnable multidimensional affine function, whereas deepspline simply requires the insertion of one more knot to a 1D CPWL function.

While it is tempting to compare architectures on the sole basis of their expressiveness, it can be very delicate to draw generic practical conclusions from this comparison. The final choice of an architecture is guided by a tradeoff between expressiveness, computation complexity, memory usage, and ability to learn over the functional space. For instance, an increase in the group size of a GroupSort activation function increases the expressiveness with no additional parameters, but usually small group sizes are favored to keep the computational impact limited.

\section{Expected Number of Regions Produced by CPWL NNs Along 1D Paths}
\label{sc:stochastic}
In Section \ref{sc:deterministic}, we found that depth increases the expressiveness of the model exponentially when the corresponding metric is the maximal number of regions. However, the compositions that achieve the lower bound of Theorem \ref{th:boundcomposition} could be very specific and hard to reach in practice.

\begin{figure}[h]
  \centering
  \centerline{\includegraphics[width=150mm]{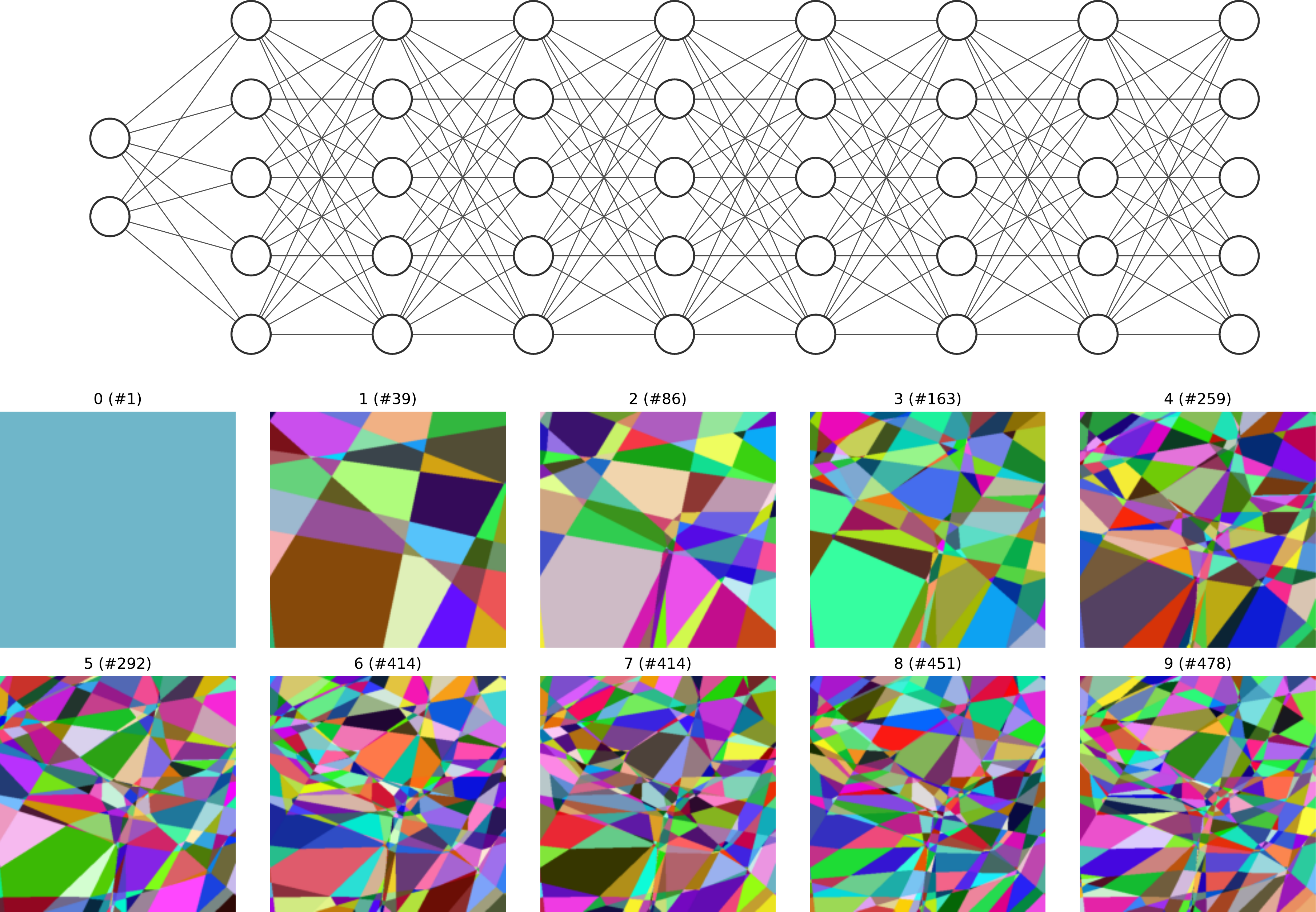}}
  \caption{Linear region-splitting process for a CPWL NN with absolute-value activation function and randomly generated parameters. The figure shows the linear regions of the mapping after $k$ activation layers, for $k=0,\ldots,9$. From one layer to the next, the regions are partitioned into smaller pieces. The number of linear regions is indicated in parentheses and suggests that the splitting process saturates with depth. The regions were numerically identified by evaluating the Jacobian of the mapping on a very fine grid.}
  \label{fig:regionsplitting}
\end{figure}

The composition $(\V f_2\circ\V f_1)$ of two CPWL functions results in the partitioning of each linear region of $\V f_1$ into smaller linear pieces. The successive compositions $(\V f_\ell\circ \cdots\circ\V f_1)$ have regions that are obtained from splitting of the regions of the previous compositions (Figure \ref{fig:regionsplitting}). As such, we expect the image of each region of the composition to shrink when depth increases, at least for compositions with reasonable gradients magnitude ($\sim 1$). The extent of the split should therefore depend on the depth of the composition. The more there are regions produced by the first compositions, the fewer splits each region will undergo after the next compositions. This intuition rules out an exponential growth of the average number of regions with $\ell$. This effect has already been revealed for ReLU NNs in \cite{Hanin2019a} and recently extended to Maxout NNs in \cite{TseranMontu2021expected}. We now aim to prove that it is universal to NNs with any type of CPWL activations under reasonable assumptions.

Throughout this section, we consider CPWL functions $\V f_{\V \theta}$ parameterized by random parameters $\V \theta$. We shall specify the parameterization and characteristics of the underlying stochastic model whenever needed. The natural extension of Section \ref{sc:deterministic} is to estimate the expected number of regions of compositions of randomly generated CPWL functions. This task seems unfortunately very complex as it mixes stochasticity and combinatorial geometry. It would involve an overly heavy framework with the risk to lose focus on the high-level intuition. Instead, we propose a simpler but closely related metric: the expected density of regions along 1D paths. This quantity is valuable in practice since it gives the expected number of linear regions that are found in-between two locations of the input space that are $1$ unit distance apart. In addition, the inverse of the density gives a rough measure of the average size of a linear region along one direction.

\subsection{Knot Density}
\change{The characterization of the density of linear regions of CPWL NNs along one-dimensional paths requires the introduction of some mathematical concepts.}
\paragraph{\change{1D CPWL Path}}A 1D CPWL path denotes in the sequel any CPWL mapping $\V\curve\colon S\rightarrow \R^d$ on a closed segment $S=[a,b]$ ($a,b\inR$) with finitely many knots. \change{This path will serve to ``navigate'' within the input domain of CPWL NNs for counting the linear regions.} The length of $\V \curve$ is computed as $\length(\V \curve) \coloneqq \int_{t\in S}\|\frac{\dint \V \curve}{\dint t}\|_2 \dint t$. Note that $\V \curve$ is a parameterization of what is often referred to as a polygonal chain. \change{In this Section we only study the density of linear regions along CPWL paths because of their simplicity and connections with CPWL NNs, e.g. the composition of a CPWL path and a CPWL NN is again a CPWL path. This choice is, however, not very restrictive since CPWL paths can approximate any continuous path arbitrarily close.}

\begin{figure}[h]
  \centering
  \centerline{\includegraphics[width=80mm]{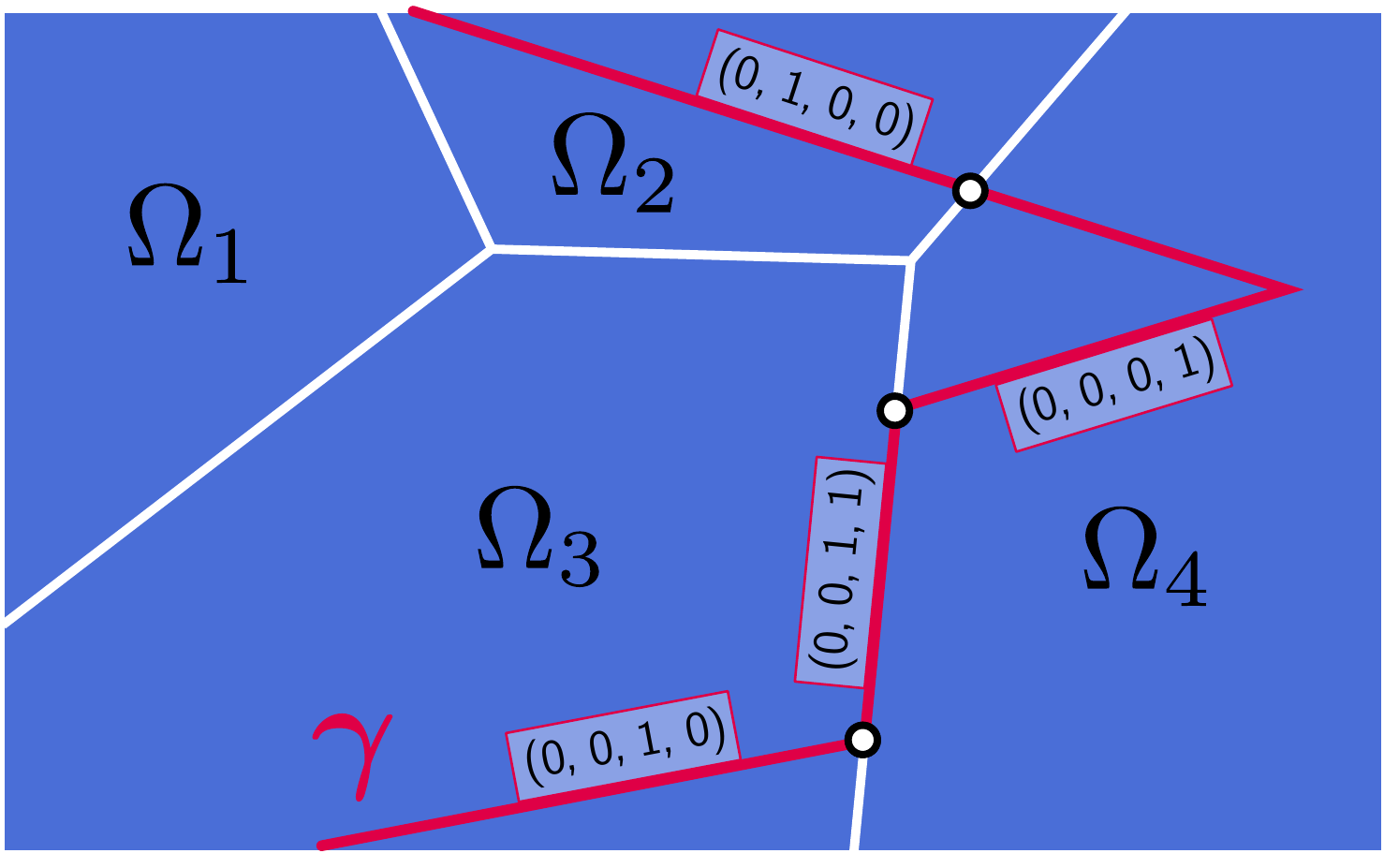}}
  \caption{\change{Example of a 1D CPWL path $\boldsymbol\gamma\colon \R \rightarrow\R^2$. The value of the characteristic function $\V \varphi_{\V f}^{\V\curve}$ along $\boldsymbol\gamma$ is given as a 4D vector and allow one to identify the 3 knots along $\boldsymbol\gamma$.}}
  \label{fig:1dcpwlpath}
\end{figure}

\paragraph{\change{Knot Density Along a Path}}
\change{Given a 1D CPWL path, the goal is to characterize the complexity of a CPWL NN along it. Informally, the number of knots of a CPWL NN along the path is the number of times the path crosses regions. This intuitive definition is unfortunately not sufficiently precise since it does not specify how} to count knots when some nonzero-length portion of the path $\V \curve$ is contained in a face of a linear region\change{, see in Figure \ref{fig:1dcpwlpath} for an example. To avoid any ambiguity, we introduce the characteristic function}
\begin{align}
  \V \varphi_{\V f}^{\V\curve}\colon \R &\rightarrow \R^{\change{K}} \\
  t & \mapsto (\mathbbm{1}_{\Omega_1}(\V\curve(t)),\ldots,\mathbbm{1}_{\Omega_K}(\V\curve(t)))
\end{align}
 of a CPWL $\V f$ along $\V \curve$, where the sets $\Omega_k$ are the projection regions of $\V f$ and $\mathbbm{1}_{\Omega_k}(\curve(t)) = 1$ if $\V \curve(t)\in\Omega_k$ and $\mathbbm{1}_{\Omega_k}(\curve(t)) = 0$ otherwise. Since $\V \curve$ is continuous with finitely many knots, and since the projection regions are unions of polyhedrons, $\varphi_{\V f}^{\V \curve}$ is a binary function with finitely many jumps\change{, see Figure \ref{fig:1dcpwlpath}}. Note that $\V \varphi_{\V f}^{\V\curve}$ uniquely identifies the supporting affine function active at location $\V\curve(t)$. Hence, in practice, the knowledge of $\V f(\V\curve(t))$ and $\nabla \V f(\V\curve(t))$, which is computable in any deep-learning library, suffice to identify $\varphi_{\V f}^{\V \curve}(t)$.

\begin{definition}[Knot density along 1D CPWL curves]
  \label{df:knotdistribution}
  Let $\V f \colon\R^{d}\rightarrow\R^{d'}$ be a CPWL function, $\V \curve$ a 1D CPWL path, and $\V\varphi^{\V\curve}_{\V f}$ the characteristic function of $\V f$ along $\V \curve$. The number $\mathrm{kt}_{\V f}^{\V \curve}$ of knots of $\V f$ along $\V \curve$ is the number of discontinuous points of the piecewise-constant function $\varphi^{\V\curve}_{\V f}$.
  The knot density $\density_{\V f}^{\V \curve}$ of $\V f$ along $\V \curve$ is defined as
  \begin{equation}
    \density_{\V f}^{\V \curve} = \mathrm{kt}_{\V f}^{\V \curve}/ \length(\V \curve),
  \end{equation}
  where $\length(\V \curve)$ is the length of $\V\curve$.
\end{definition}

We stress that alternative definitions of the knot density that correspond to the same informal intuition are possible, but they would differ when the path $\V\curve$ follows the boundaries of some projection regions. In the sequel, this will not matter since, in any reasonable stochastic framework, the path does not follow some boundaries almost surely.

The knot density along a path is subadditive for the sum and vectorization of CPWL functions\changer{, and can be bounded for the composition of CPWL functions, see Proposition \ref{pr:densitysumveccompo} and \ref{pr:compolinear} and \ref{ap:density_bounds_operation} for the corresponding proofs.}
\begin{proposition}
  \label{pr:densitysumveccompo}
  Let $\V \curve\colon S\rightarrow \R^{d}$ be a 1D CPWL path on the segment $S\subset\R$ and let $\V f_1\colon\R^{d}\rightarrow\R^{d'}$ and $\V f_2\colon\R^{d}\rightarrow\R^{d'}$ be two CPWL functions. The knot density along $\V \curve$ of either the sum $\V f_1+\V f_2$ or of the vectorized function $(\V f_1,\V f_2)$ is bounded as
  \begin{align}
    \lambda_{\V f_1+\V f_2}^{\V \curve}&\leq \density_{\V f_1}^{\V \curve}+\density_{\V f_2}^{\V \curve},\\
    \density_{(\V f_1,\V f_2)}^{\V \curve}&\leq\density_{\V f_1}^{\V \curve}+\density_{\V f_2}^{\V \curve},
  \end{align}
  where $\density_{\V f_1}^{\V \curve}$ and $\density_{\V f_2}^{\V \curve}$ are the knot density of $\V f_1$ and $\V f_2$ along $\V\curve$, respectively.
\end{proposition}

\begin{proposition}
  \label{pr:compolinear}
  Let $\V \curve\colon S\rightarrow \R^{d_1}$ be a 1D CPWL path on $S\subset\R$ and let $\V f_1\colon\R^{d_1}\rightarrow\R^{d_2}$ and $\V f_2\colon\R^{d_1}\rightarrow\R^{d_2}$ be two CPWL functions. Then, the knot density of $\V f_2\circ\V f_1$ on $\V \curve$ is bounded as
  \begin{equation}
    \density_{\V f_2\circ \V f_1}^{\V \curve} \leq \density_{\V f_1}^{\V \curve} + \left(\frac{\length{(\V f_1\circ \V \curve})}{\length{(\V \curve)}}\right)\density_{\V f_2}^{\V f_1\circ \V \curve},
  \end{equation}
  where $\density_1^{\V \curve}$ is the knot density of $\V f_1$ along $\V\curve$ and $\density_2^{\V \curve}$ the one of $\V f_2$ along $\V f_1\circ\V \curve$.
\end{proposition}

\subsection{Knot Density of CPWL Layers}
\label{subsec:densityCPWLnets}
The goal of this subsection is to show that the knot density is well behaved for classical CPWL NN layers, which justifies the assumption $(i)$ of Theorem \ref{th:boundstochastic} and Corollary \ref{cr:boundstochastic}. The proofs can be found in \ref{ap:densitybounds}.
\begin{proposition}[Knot density - ReLU]
  \label{pr:knotdensityReLU}
  Let $(\V w,b)\in\R^d\times\R$ be independent random variables with bounded probability density functions $\rho_{b}$ for b and $\rho_{w}$ for the components of $\V w$, which are i.i.d. Then, the expected knot density of the ReLU CPWL component $\M x\mapsto \mathrm{ReLU}(\V w^T \M x + b)$ along any 1D CPWL path $\V \curve$ is bounded as
  \begin{equation}
    \mathbb{E}\left[\density_{\mathrm{ReLU}}^{\V \curve}\right] \leq \sqrt{\mathbb{E}[w_1^2]} \sup_{t\inR}\rho_b(t).
  \end{equation}
 In particular, when $b$ and the components of $\V w$ are normally distributed with zero mean and standard deviation $\sigma_b$ and $\sigma_w$, respectively, the following tighter bound holds true
  \begin{equation}
    \mathbb{E}\left[\density_{\mathrm{ReLU}}^{\V \curve}\right] \leq \frac{\sigma_w}{\pi\sigma_b}.
  \end{equation}
\end{proposition}
When the ReLU activation function is replaced by a 1D CPWL function with a given number $K$ of knots, we conjecture that the bounds can simply be multiplied by $K$.

\begin{proposition}[Knot density - Maxout]
  \label{pr:knotdensityMaxout}
Let $((w_{k1},\ldots,w_{kd}),b_k)\inR^d\times\R$ for $k=1,\ldots,K$ be independent random variables with bounded probability density functions $\rho_{b}$ for any $b_k$ and $\rho_w$ for all components $w_{kl}$ of $\V w_k$, which are i.i.d. over both $k\in[K]$ and $l\in[d]$. Then, the expected knot density of the rank $K$ Maxout unit $f\colon \M x\mapsto \max_{k=1,\ldots,K}(\V w_k^T \M x + b_k)$ along any 1D CPWL path $\V \curve$ is bounded as
  \begin{equation}
    \mathbb{E}\left[\density_{\mathrm{Maxout}}^{\V \curve}\right] \leq \sqrt{2}{K\choose 2}\sigma_w\sup_{t\inR}\rho_b(t),
  \end{equation}
  where $\sigma_w$ is the standard deviation of any $w_{kl}$.
  In particular, when $b_k$ and $w_{kl}$ are normally distributed with zero mean and standard deviation $\sigma_b$ and $\sigma_w$, respectively, a tighter bound holds true, according to
  \begin{equation}
    \mathbb{E}\left[\density_{\mathrm{Maxout}}^{\V \curve}\right] \leq \sqrt{2}{K\choose 2}\frac{\sigma_w}{\pi\sigma_b}.
  \end{equation}
\end{proposition}

The bounds provided in Proposition \ref{pr:knotdensityMaxout} grow quadratically in terms of the number of Maxout units; we conjecture the existence of a tighter linear bound. 
\begin{proposition}[Knot density - GroupSort]
  \label{pr:knotdensityGS}
  Let $(\V w_{k},b_k)$ be as in Proposition \ref{pr:knotdensityMaxout}. Then, the expected knot density $\density_{\V f}^{\V \curve}$ of the GroupSort layer $\V f\colon \R^d\rightarrow\R^d\colon \M x\mapsto \mathrm{GS}_{n_g,g_s}(\V W \M x)$, where $\mathrm{GS}_{n_g,g_s}$ is the GroupSort activation with $n_g$ groups of size $g_s$, is bounded along any 1D CPWL path $\V \curve$ as
  \begin{equation}
    \mathbb{E}\left[\density_{\mathrm{GroupSort}}^{\V \curve}\right] \leq \frac{\sqrt{2}}{2} d (g_s-1) \sigma_w\sup_{t\inR}\rho_b(t),
  \end{equation}
  where $\sigma_w$ is the standard deviation of any $w_{k,l}$.
  In particular, when $b_k$ and $w_{kl}$ are normally distributed with zero mean and standard deviation $\sigma_b$ and $\sigma_w$, respectively, a tighter bound can be given as
  \begin{equation}
    \mathbb{E}\left[\density_{\mathrm{GroupSort}}^{\V \curve}\right] \leq \frac{\sqrt{2}}{2} d (g_s-1) \frac{\sigma_w}{\pi\sigma_b}.
  \end{equation}
\end{proposition}

For ReLU and Maxout layers with multidimensional outputs, the bounds given in Proposition \ref{pr:knotdensityReLU} and \ref{pr:knotdensityMaxout} are simply multiplied by the output dimension (see Proposition \ref{pr:densitysumveccompo}).
We note that all bounds proposed take the form $(\kappa W\sigma_w\sup_{t\inR}\rho_b(t))$, where the prefactor $\kappa$ only depends on the activation function and $W$ is the number of outputs of the layer. The learnable parameters are typically initialized by sampling a uniform or normal distribution with the same characteristics for the biases and the weights of a same layer. In this case, although the characteristics of the distribution usually depend on the input and output dimensions of the layer \cite{He2015}, the quantity $\sigma_w\sup_{t\inR}\rho_b(t)$ is determined only by the distribution: normal or uniform (since, for these distributions, the supremum of the probability density function is proportional to the standard deviation).
All in all, it should be reminded that
\begin{itemize}
  \item the expected knot density is well defined for learnable CPWL layers;
  \item with standard initialization methods, it is reasonable to assume that the expected knot density of the components of a CPWL layer depends neither on its width nor on the total depth of the NN (at least at initialization stage).
\end{itemize}
It is tempting to take advantage of the previous results to adjust the distributions of the weights and biases at initialization in the hope to increase the upper bound and, possibly, the knot density of a NN. The effect is, however, subtle: for instance, if one narrows the distribution of the biases, the bound increases as $\sup_{t\inR}\rho_b(t)$ increases. While this may increase the average knot density at some specific locations, it will inevitably decrease it elsewhere.

\subsection{Bounds on the Expected Knot Density of CPWL NNs}
In Theorem \ref{th:boundstochastic} and Corollary \ref{cr:boundstochastic}, we introduce two different settings to bound the expected knot density of CPWL NNs. Theorem \ref{th:boundstochastic} highlights the role played by the gradients of the composed layers: larger gradients allow for a more intense splitting process within the composition and should lead to a greater knot density. With Corollary \ref{cr:boundstochastic}, we propose a more practical analysis: given a learning task that dictates the input and output dimensions, how does the expected density of linear regions along 1D curves relate to the depth, width, and activation complexity of the CPWL NN? In accordance with the intuition given in Figure \ref{fig:regionsplitting}, depth cannot provide exponentially more linear regions on average. This key result relies mainly on the assumption $(ii)$, which is discussed in Section \ref{subsec:discussionstochabound}.

The directional derivative of the function $\V f$ in the direction $\M u$ is denoted by $D_{\M u}\{\V f\}$\changer{, and the proofs of the results proposed in this section can be found in \ref{ap:density_cpwl_nn}.}
\begin{theorem}
  \label{th:boundstochastic}
  Let $\V f_{\V \theta_1},\ldots,\V f_{\V \theta_L}$, with $\V f_{\V \theta_\ell}\colon\R^W\rightarrow\R^W$, be CPWL functions parameterized by the independent and identically distributed random variables $\V \theta_1,\ldots,\V \theta_L$. Suppose that there exist $\density_0,D_0\inR$ such that
  \begin{enumerate}[label=(\roman*)]
    \item for any 1D CPWL path $\V \curve$, $\mathbb{E}[\density_{f_{\V \theta_\ell,k}}^{\V \curve}]\leq \density_0$, where $f_{\V \theta_\ell,k}$ is the $k$th component of $\V f_{\V \theta_\ell}$ (bounded expected knot density of the components);
    \item for any $\M x, \M u \inR^W$ with $\|\M u\|_2=1$, $\mathbb{E}[D_{\M u}\{\V f_{\V \theta}\}(\M x)]\leq D_0$ (bounded expected directional derivative).
  \end{enumerate}
  Then, on any 1D CPWL path $\V \curve$, the expected knot density of the CPWL NN is bounded as
  \begin{equation}
    \mathbb{E}[\density_{\V f_{\V \theta_L}\circ\cdots\circ \V f_{\V \theta_1}}^{\V \curve}]\leq
    \begin{cases}
    \density_0 W \left(\frac{1 - D_0^L}{1 - D_0} \right), & D_0\neq 1 \\
    \density_0 W L, & D_0 = 1.
    \end{cases}
    \end{equation}
\end{theorem}
\begin{corollary}
  \label{cr:boundstochastic}
  Let $\V f_{\V \theta_1},\ldots,\V f_{\V \theta_L}$, with $\V f_{\V \theta_\ell}\colon \R^{d_\ell}\rightarrow\R^{d_{\ell+1}}$, be CPWL functions parameterized by the independent and identically distributed random variables $\V \theta_1,\ldots,\V \theta_L$ and $d_2=\cdots=d_L=W>d_{L+1}$. Suppose that there exist $\density_0,D_0\inR$ such that
  \begin{enumerate}[label=(\roman*)]
    \item for any 1D CPWL path $\V \curve$, $\mathbb{E}[\density_{f_{\V \theta_\ell,k}}^{\V \curve}]\leq \density_0$, where $f_{\V \theta_\ell,k}$ is the $k$th component of $\V f_{\V \theta_\ell}$ (bounded expected knot density of the components),
    \item for any $\M x, \M u \inR^d$ with $\|\M u\|_2=1$, $\mathbb{E}[D_{\M u}\{\V f_{\V \theta_\ell}\circ\cdots\circ \V f_{\V \theta_1}\}(\M x)]\leq D_0$, for $1\leq \ell\leq L$ (bounded expected directional derivative within the composition).
  \end{enumerate}
  Then, on any 1D CPWL path $\V \curve$, the expected knot density of the CPWL NN is bounded as
  \begin{equation}
    \mathbb{E}[\density_{\V f_{\V \theta_L}\circ\cdots\circ \V f_{\V \theta_1}}^{\V \curve}]
    \leq D_0^* (\density_0 W L),
    \end{equation}
where $D_0^*=\max(D_0,1)$.
\end{corollary}
The proof of Theorem \ref{th:boundstochastic} relies on Lemma \ref{lm:compolength}. In the bound presented in this lemma, the expected value is evaluated before taking the supremum, whilst a switch of the order of the operators would yield a much looser bound. 
\begin{lemma}
  \label{lm:compolength}
  Let $\V \curve\colon S\rightarrow \R^d$ be a 1D CPWL path and $\V f_{\V \theta}\colon \R^d\rightarrow\R^{d'}$ a CPWL function parameterized by the random variable $\V \theta$ such that, for any $\M x,\M u \inR^d$, $\V f_{\V \theta}$ is differentiable at $\M x$ in direction $\M u$ with probability 1. Then, the expected length of the 1D CPWL path $\V f_{\V \theta}\circ\V\curve\colon S\rightarrow \R^{d'}$ is bounded as
  \begin{equation}
    \mathbb{E}[\length(\V f_{\V \theta}\circ\V\curve)]\leq \length(\V \curve) \sup_{\substack{\M x,\M u \inR^d\\ \|\M u\|_2 = 1}}\mathbb{E}[\|D_{\M u}\{\V f_{\V \theta}\}(\M x)\|_2].
  \end{equation}
\end{lemma}

\subsubsection{Discussion of the Compositional Bounds}
\label{subsec:discussionstochabound}
Our approach relies on the independence of the randomly generated CPWL functions. It usually holds at initialization stage, but it is not true anymore in the learning stage. While this can be regarded as a limitation, it is a legitimate and convenient way to explore and depict the whole function space that a given architecture gives access to.

Assumption $(i)$ of Theorem \ref{th:boundstochastic} and its corollary (bounded expected knot density of the learnable CPWL components) have been discussed in details in Section \ref{subsec:densityCPWLnets}, where it was remarked that it is reasonable to assume that $\lambda_0$ is independent of $W$ and $L$.

Theorem \ref{th:boundstochastic} and Corollary \ref{cr:boundstochastic} differ on Assumption $(ii)$ (well behaved gradients). While the assumption of the theorem seems more natural at first sight (gradient controlled for each layer), the one of the corollary is closer to practical observations. Assumption $(ii)$ of Corollary \ref{cr:boundstochastic} was invoked to bound the expected length of the image of any finite-length 1D CPWL path, independently of the depth of the composition. \change{While early works suggested that this expected length grows exponentially with depth \cite{pmlr-v70-raghu17a}, it was recently shown otherwise in a more realistic setup, both theoretically and experimentally \cite{hanin2021}}. For instance, for ReLU NNs, with the usual 2/fan-in weight variance, depth typically does not affect the expected length \cite{hanin2021}. More generally, a control of the magnitude of the directional derivatives that is independent of the depth is highly desirable in the learning stage for a stable back-propagation algorithm \cite{He2015} and, in the inference stage, to produce robust models \cite{Pauli2022}. In short, it is also reasonable to assume that the parameter $D_0$ depends neither on $W$ nor on $L$.

The previous discussion suggests a simple and important bound on the density of regions of CPWL NNs. It attributes an identical role to the three sources of complexity, namely depth, width, and activation complexity.

The quality of the proposed bounds seems to be completely determined by the tightness of the bounds in Assumptions $(i)$ and $(ii)$. Based on the proofs of Theorem \ref{th:boundstochastic} and Corollary \ref{cr:boundstochastic}, we believe that the compositional bounds are sharp provided that the expected knot density is uniform (i.e., the same for any 1D CPWL curve) and that the expected norm of the directional derivative is uniform and isotropic within the NN.

\section{Conclusion}
In this work, we have investigated the role of depth, width, and activation complexity in the expressiveness of CPWL NNs. By invoking results from combinatorial geometry, we have found that depth has a predominant role over width and activation complexity: it is the only descriptor able to increase the number of linear regions exponentially. However, this exponential growth is only observed for the maximal number of regions. Indeed, when exploring the whole function space produced by a given CPWL NN, we have found that, on average, the number of regions along a line is bounded by a quantity that only depends on the product of the three descriptors. In that perspective, the three complexity parameters have an identical role: no exponential behavior with depth is observed anymore.

The ability to train deeper and deeper NNs has led to major improvements in machine learning. However, depth comes at a price in applications where the NN needs to be stable, for instance by constraining its global Lipschitz constant. In such settings, we therefore believe that complex learnable activations should always be regarded as a valuable opportunity to increase substantially the expressiveness of the model without resorting to deeper NNs \cite{bohra2021learning, Neumayer22Lip}.
\subsection*{Acknowledgments}
The authors are thankful to Shayan Aziznejad for helpful discussions.

\vskip 0.2in
\bibliography{All}

\newpage
\appendix
\addcontentsline{toc}{section}{Appendices}
\section*{Appendices}
\section{Proofs for Section \ref{sc:deterministic}}
\label{ap:proofsec2}
\subsection{Number of Convex vs Projection Regions}

\begin{proof}{\bf of Proposition \ref{pr:projectionvsconvex}}
  The first inequality follows from the fact that there cannot be fewer linear convex regions than affine pieces.
  Consider two neighboring projection regions $\Omega_k$ and $\Omega_\change{p}$ of $\V f$, where $1\leq k< p\leq \rho$,  with corresponding affine pieces $\V f^k\colon \M x\mapsto \M W_k^T\M x + \M b_k$ and $\V f^p\colon \M x\mapsto \M W_p^T\M x + \M b_p$, where $\M W_k,\M W_p \inR^{d'\times d}$ and $\M b_k,\M b_p\inR^{d'}$. Since $\V f$ is continuous, any $\M x \in \Omega_k\cap \Omega_p$ satisfies that $(\M W_k-\M W_p)^T\M x + (\M b_k - \M b_p)=0$. The set of all boundary points of $\V f$ is therefore included in $\cup_{1\leq k< p\leq \rho} H_{kp}$, where $H_{kp}=\{\M x\in\R^d\colon (\M W_k-\M W_p)^T\M x + (\M b_k - \M b_p)=0\}$ is an affine subspace of dimension at most $(d-1)$ since $k\neq p$. The arrangement of the $\rho(\rho-1)/2$ hyperplanes $H_{kq}$ with $k\neq \change{p} \in [\rho]$ yields convex regions on which $\V f$ is affine since these regions do not contain boundary points. The number of such regions is, therefore, an upper bound on the number of convex regions of $\V f$. \change{It is known from \cite{zaslavsky1975facing} that the number of convex regions formed by an arrangement of $N$ hyperplanes in $\R^d$ is at most $\sum_{k=0}^{\min(d,N)}{N\choose k}$. Hence, for $\rho(\rho-1)/2>d$, we directly reach the announced result. Otherwise, the bound yields $\sum_{k=0}^{\rho(\rho-1)/2}{\rho(\rho-1)/2\choose k}=2^{\rho(\rho-1)/2}$}.
\end{proof}

  \begin{proof}{\bf of Lemma \ref{lm:subpartition}}
 Let $e = \mathrm{dim} (E)$. The natural candidate for $\partition_E$ is the partition
  \begin{equation}
  \partition'=\left \{ P'\colon P' = P \cap E,  P \in \partition, \hbox{ and } \mathrm{Int} P' \neq \emptyset \right \},
  \end{equation}
  which is unfortunately not necessarily a proper convex partition. Indeed, if $E$ contains an $e$-face of a region, then some elements of $\partition'$ will not have disjoint interiors. Since the regions of $\partition$ are polyhedrons, there exist a given number $n_H$ of distinct boundary hyperplanes $H_p=\{\M x\inR^d\colon \mathbf{a}_{p}^T \mathbf{x}+b_{p} = 0\}$ and such that for each $P_k \in \partition$, there exists a subset $I_k\subset [n_H]$ and $\epsilon_{k,p}\in\{-1,1\}$ for $p\in I_k$ such that
  \begin{equation}
    P_k = \{\mathbf{x}\in \mathbb{R}^d \colon \epsilon_{k,p}(\mathbf{a}_{p}^T \mathbf{x}+b_{p}) \geq 0\quad\forall p\in I_k\}.
    \end{equation}
    We now consider a mapping $\phi$ that assigns to each hyperplane $H_p$ a unique region $\phi(p)$ such that $p\in I_{\phi(p)}$. We can now define $n$ new pairwise-disjoint convex regions as
    \begin{equation}
      P'_k = \left\{\mathbf{x}\in \mathbb{R}^d \colon \begin{matrix} \epsilon_{k,p}(\mathbf{a}_{p}^T \mathbf{x}+b_{p}) \geq 0 & \text{for } p\in[n_k] \text{ and }\phi(p) = k\\
        \epsilon_{k,p}(\mathbf{a}_{p}^T \mathbf{x}+b_{p}) > 0 & \text{for } p\in[n_k] \text{ and } \phi(p) \neq k
      \end{matrix}\right\}.
      \end{equation}
      It is clear that $\cup_{k=1}^n P'_k = \R^d$.
From these new regions, one can eventually build the proper convex partition
\begin{equation}
  \partition_E=\left \{ P_E = \overline{P'_k \cap E} \colon k\in[n] \hbox{ and } \mathrm{Int} P_E \neq \emptyset\right \}.
  \end{equation}
By construction, all regions of $\partition_E$ are closed with nonempty interiors; their union covers $E$. Let $P_{E,1}=P_{k_1}'\cap E$ and $P_{E,2}=P_{k_2}'\cap E$ be two (nonempty) regions of $\partition_E$. We have that $\mathrm{Int }(P_{E,1})\cap \mathrm{Int }(P_{E,2}) = \mathrm{Int }(P_{E,1}\cap P_{E,2}) = \mathrm{Int }(P_{k_1}'\cap P_{k_2}'\cap E)=\emptyset$ for $k_1\neq k_2$. We, therefore, proved that $\partition_E$ is a convex partition of $E$; it has at most $n$ regions and is such that, for any $P_E\in \partition_E$, there is $P\in \partition$ with $P_E \subset P$.
  \end{proof}

  \begin{proof}{\bf of Lemma {\ref{lm:inverseimagepartition}}}
  Let $P \in \partition$. Recall that $P$ is a closed and convex subset of the affine space $\V f(\mathbb{R}^d)$ with dimension $\mathrm{dim}(\V f(\R^d))$. We first prove that $\V f^{-1}(P)$ meets the requirements to form a convex partition of $\R^d$.
  \begin{itemize}
    \item The continuity of $\V f$ implies that $\V f^{-1}(P)$ is closed.
    \item The function $\V f$ is written as $\V f\colon \M x \mapsto \M A \M x +\M b$ with $\M A\inR^{d'\times d}$ and $\M b\inR^{d'}$. For $\M x, \M y \in \V f^{-1}(P)$ and $\ub \in [0,1]$, we have that $\V f(\ub \M x + (1-\ub) \M y) =\M A ( \ub \M x + (1-\ub)\M y) + \M b = \ub \V f(\M x) + (1-\ub) \V f(\M y)\in P$ since $P$ is convex. Therefore, $\V f^{-1}(P)$ is also convex.
    \item We have that $\cup_{P\in \partition} \V f^{-1}(P) = \V f^{-1}(\cup_{P\in \partition}P) =  \V f^{-1}(\V f(\R^d)) = \R^d$.
    \item For two distinct regions $P_1,P_2\in\partition$, we have that $\V f^{-1}(P_1)\cap \V f^{-1}(P_2)=\V f^{-1}(P_1\cap P_2)$. Since $P_1$ and $P_2$ are distinct regions of $\partition$, $\mathrm{dim}(P_1\cap P_2)< \mathrm{dim}(\V f(\R^d))$, which implies that $\mathrm{dim}(\V f^{-1}(P_1)\cap \V f^{-1}(P_2))< d$ and proves that $P_1$ and $P_2$ have disjoint interiors.
    \item We decompose the input space as the direct sum $\R^d=\mathrm{ker}(\M A)\oplus U$. Note that $\V f(U) = \V f(\R^d)$. It is clear that, for any $\M x\in \V f^{-1}(P)$ and $\M y\in \mathrm{ker}(\M A)$, we have that $\M x +\M y\in \V f^{-1}(P)$, which implies that $\mathrm{dim}(\mathrm{Proj}_U(\V f^{-1}(P)))=\mathrm{ker}(\M A)$. In addition, we use the fact that $\V f$ restricted to $U$ has full rank and write $\mathrm{dim}(\mathrm{Proj}_{\mathrm{ker}(\M A)}(\V f^{-1}(P))) = \mathrm{dim}(P)=\mathrm{dim}(\V f(\R^d))$. All in all, we have proved that $\mathrm{dim}(\V f^{-1}(P))=d$, which implies that the regions of $\V f^{-1}(\partition)$ have nonempty interiors.
 \end{itemize}
  \end{proof}

    \begin{proof}{\bf of Lemma \ref{lm:dim}}
      The result stems from the fact that the rank of a product of matrices is bounded by the smallest rank of these matrices.
    \end{proof}

\subsection{Upper Bound on the Number of Regions of Arrangements}

\begin{proof}{\bf of Theorem \ref{th:upperbound}}
  First, we prove that the expression given in the theorem is an upper bound. To that end, we need to formalize our problem with the notion of abstract simplicial complex so as to focus solely on the combinatorial structure of the task and be compliant with the formalism of \cite{Bulavka2020}. Let $\partition^*_k=\{\mathrm{int}(P)\colon P\in \partition_k\}$, where $\mathrm{int}(P)$ denotes the interior of $P$ in $\R^d$, and let $\mathcal{F} = \cup_{k=1}^N \{P^*\colon P^*\in \partition^*_k\}$ be the set that contains the elements of the $N$ sets $\partition^*_k$. The the nerve $\mathrm{K}$ of $\mathcal{F}$ is defined as
    \begin{equation}
    \mathrm{K} = \{ X \subset \mathcal{F} \colon \cap X \neq \emptyset \}.
    \end{equation}
    In simple words, $\mathrm{K}$ is made of all the nonempty intersections of sets in any of the $\partition^{*}_k$.
    The nerve of an open covering is an abstract simplicial complex which, therefore, applies to $\mathrm{K}$ since $\mathcal{F}$ is an open covering of $\R^d$. This more simply follows from the definition of an abstract simplicial complex: it is a family of sets that is closed under taking subsets. In the sequel, we need $\mathrm{K}$ to be a $d$-representable simplicial complex, which is granted because it is the nerve of a finite family of convex sets in $\mathbb{R}^d$ (more details in \cite{Bulavka2020}). In our problem, the faces of dimension 0 of the complex, also known as vertices, are the elements of $\mathcal{F}$. More generally, a face of $\mathrm{K}$ of dimension $p$ is a nonempty intersection of $p+1$ elements of $\mathcal{F}$. Each set $\partition^*_k$ induces a sub-complex $\mathrm{K}[\partition^*_k]=\{ X \subset \partition^*_k \colon \cap X \neq \emptyset \}$ of $\mathrm{K}$. The dimension of this sub-complex, which is the largest dimension of its faces, is 0 because the elements of $\partition^*_k$ are disjoint. We note that the interior of the regions of the arrangement of the convex partitions are $(N+1)$-faces of the abstract simplicial complex $\mathrm{K}$, which are also called $\M 1$-colorful faces, where $\M 1=(1,\ldots,1)\inR^N$ specifies that each region of the arrangement is built from one region per partition. We are therefore looking to bound the number $f_{\mathbf{1}}(\mathrm{K})$ of $\M 1$-colorful faces of the complex $\mathrm{K}$. Since we have now fully translated our problem into the framework of \cite{Bulavka2020}, we can apply \cite[Theorem 10]{Bulavka2020} to $\mathcal{F}$. The parameter $\mathbf{r}=(r_1,\ldots,r_N)$ can be chosen so that $\mathrm{dim}(\mathrm{K}[\partition^*_k])\leq (r_k - 1)$. Therefore, we simply choose $\M r = \M 1$ and obtain that
    \begin{equation}
    \ub^d(n_1,\ldots,n_N) = f_{\mathbf{1}}(\mathrm{K}) \leq p_{\mathbf{1}}(\mathbf{n},d,\mathbf{1}),
    \end{equation}
    where 
    \begin{equation}
    p_{\mathbf{k}}(\mathbf{n},d,\mathbf{r}) =\sum_{\mathbf{\ell}=(\ell_1, \dots, \ell_{N}) \in L_{\mathbf{k}}(d)} \prod_{i=1}^{N} \binom{n_i-r_i}{\ell_i}\binom{r_i}{k_i-\ell_i}
    \end{equation}
    and
    \begin{equation}
    \label{e:L_kd}
      L_\mathbf{k}(d) = \{\mathbf{\ell} = (\ell_1, \ldots \ell_N) \in \mathbb{N}^{N} \colon
      \ell_1 + \cdots + \ell_N \leq d \hbox{ and }\ell_i \leq
      k_i \hbox{ for }i \in[N]\}.
    \end{equation}
In our problem, $\M k=\M 1$ and
    \begin{equation}
      L_\mathbf{1}(d) = \{\mathbf{\ell} = (\ell_1, \ldots \ell_N) \in \mathbb{N}^{N} \colon \ell_1 + \cdots + \ell_N \leq d \hbox{ and }
      \ell_i \in \{0,1\} \hbox{ for }i \in[N]\}.
    \end{equation}
    With $\M r=\M 1$, we have that
    \begin{align}
      \nonumber
    \ub^d(n_1,\ldots,n_N) \leq p_{\mathbf{1}}(\mathbf{n},d,\mathbf{1}) &=\sum_{\mathbf{\ell}=(\ell_1, \dots, \ell_{N}) \in L_{\mathbf{1}}(d)} \prod_{i=1}^{N} \binom{n_i-1}{\ell_i}\binom{1}{1-\ell_i}\\
    \nonumber
    &= \sum_{\mathbf{\ell}=(\ell_1, \dots, \ell_{N}) \in L_{\mathbf{1}}(d)} \prod_{i=1}^{N} \binom{n_i-1}{\ell_i}\\
    \nonumber
    &= \sum_{k=0}^{d}\sum_{\substack{ \ell_1,\ldots,\ell_{N}\in\{0,1\}\\
    \nonumber\ell_1+\cdots+\ell_{N} = k}} \prod_{i=1}^{N} \binom{n_i-1}{\ell_i}\\
    \nonumber
    &= \sum_{k=0}^{d}\sum_{\substack{ \ell_1,\ldots,\ell_{N}\in\{0,1\}\\
    \nonumber\ell_1+\cdots+\ell_{N} = k}} \prod_{\substack{i=1\\\ell_i=1}}^{N} \binom{n_i-1}{\ell_i} \prod_{\substack{i=1\\
    \nonumber
    \ell_i=0}}^{N} \binom{n_i-r_i}{\ell_i}\\
    &= \sum_{k=0}^{d}\sum_{\substack{ \ell_1,\ldots,\ell_{N}\in\{0,1\}\\
    \nonumber
    \ell_1+\cdots+\ell_{N} = k}} \prod_{\substack{i=1\\
    \nonumber
    \ell_i=1}}^{N} (n_i-1)\\
    &=1 + \sum_{k=1}^d \sum_{1\leq \ell_1<\cdots<\ell_k\leq N} \prod_{i=1}^k(n_{\ell_i}-1),
    \end{align}
    which proves that the bound given in the Theorem holds true.

    Now we show that this upper bound is sharp. To that end, consider that each partition $\partition_k$ is made of the regions of the arrangement of the $(n_k-1)$ distinct parallel hyperplanes $H_q^k$ for $q=1,\ldots,(n_k-1)$ so that the hyperplanes are in general position when only one per partition is selected. Recall that $N$ hyperplanes are in general position if any collection of $k$ of them intersect in a $(d-k)$-dimensional plane for $1\leq k \leq d$ and have empty intersection for $k > d$. The number of regions of the arrangement $\mathcal{A}(\partition_1,\ldots,\partition_N)$ is exactly the number of regions of the arrangement of all the hyperplanes $H_q^k$ for $q=1,\ldots,(n_k-1)$ and $k=1,\ldots,N$. Following Zavalasky's Theorem, the number of regions can be computed by 
    \begin{equation}
    \#\mathcal{R}(\mathcal{A}) = (-1)^d \chi_{\mathcal{A}}(-1),\end{equation}
   where $\chi_{\mathcal{A}}$ is the characteristic polynomial of the arrangement. There is no need here to define the characteristic polynomial in detail since Whitney's formula provides a direct way to evaluate it as
  \begin{equation}
     \chi_{\mathcal{A}}(-1) = \sum_{\substack{\mathcal{B}\subset\mathcal{A} \\ \cap_{H \in \mathcal{B}}H\neq \emptyset}}(-1)^{\# \mathcal{B}}(-1)^{\mathrm{dim}( \cap_{H \in \mathcal{B}}H)
    }.
  \end{equation}
    The subsets $\mathcal{B}\subset\mathcal{A}$ that have a nonempty intersection can be written as $\mathcal{B} =\{H_{q_{k_1}}^{k_1},\ldots,H_{q_{k_p}}^{k_p}\}$ with $1\leq k_1<\cdots< k_p \leq N$, $q_{k_i}\in [n_{k_i}-1]$ where $i=1,\ldots,p$ and $0\leq p\leq d$. This holds because, for $q\neq q'$, $H_q^k\cap H_{q'}^k = \emptyset$. Note that, by convention, the set $\mathcal{B} = \emptyset$ is also considered in the sum. Because of the particular choice of the hyperplanes, for a given $p$, $\mathcal{B}$ is the nonempty intersection of $p$ hyperplanes and there are $\sum_{1\leq\ell_1<\cdots<\ell_p \leq N}\prod_{i=1}^p(n_{\ell_i}-1)$ such subsets of $\mathcal{A}$. The intersection of the elements of $\mathcal{B}$ has dimension $(d-p)$ (recall that the hyperplanes of $\mathcal{B}$ are in general position). All in all, we have that
    \begin{align}
      \nonumber
    \#\mathcal{R}(\mathcal{A}) &= (-1)^d \sum_{\substack{\mathcal{B}\subset\mathcal{A}\colon \\ \nonumber
    \cap_{H \in \mathcal{B}}H\neq \emptyset}}(-1)^{\# \mathcal{B}}(-1)^{\mathrm{dim}( \cap_{H \in \mathcal{B}}H)
    }\\
    \nonumber
    &= 1 + (-1)^d \sum_{k=1}^d \sum_{ 1\leq\ell_1<\cdots<\ell_k \leq N}(-1)^{k}(-1)^{d-k}\prod_{i=1}^k (n_{\ell_i}-1) \\
    &= 1 + \sum_{k=1}^d \sum_{ 1\leq\ell_1<\cdots<\ell_k \leq N}\prod_{i=1}^k (n_{\ell_i}-1),
    \end{align}
    which is the upper bound given in the theorem.

    When $N\leq d$, we readily check that the bound is giving $n_1\cdots n_N$. To prove the second additional bound for $N>d$ given in the theorem, we invoke the binomial theorem and remark that
    \begin{align}
      \nonumber
      \left(1+\sum_{p=1}^N (n_p-1)\right)^d &= 1+ \sum_{k=1}^d {d\choose k}\left(\sum_{p=1}^N (n_p-1)\right)^k= 1+ \sum_{k=1}^d {d\choose k}\sum_{1\leq\ell_1,\ldots,\ell_k \leq N}\prod_{i=1}^k(n_{\ell_i}-1)\\
      \nonumber
      &\geq  1+ \sum_{k=1}^d \sum_{1\leq\ell_1,\ldots,\ell_k \leq N}\prod_{i=1}^k(n_{\ell_i}-1)\\
      &\geq 1+ \sum_{k=1}^d \sum_{1\leq\ell_1<\cdots<\ell_k \leq N}\prod_{i=1}^k(n_{\ell_i}-1).
    \end{align}
    \end{proof}

\subsection{Sum and Vectorization}

\begin{proof}{\bf of Proposition \ref{pr:sumandvectorization}}
  Let $\partition_k$ be a linear convex partition of $f_k$ for $k=1,\ldots,N$. On each region of the arrangement $\mathcal{A}(\partition_1,\ldots,\partition_N)$, the $f_k$ are affine, and so is their sum and their vectorization. This implies that $\mathcal{A}(\partition_1,\ldots,\partition_N)$ is a linear convex partition of both the sum and the vectorization of the scalar-valued CPWL functions, which shows that $\ub^d(\reg_1,\ldots,\reg_N)$ is a valid upper bound on the number of convex linear regions.
  
  We now prove that the bounds are sharp. First, consider $N$ convex partitions $\partition_k$ where each $\partition_k$ is made of the regions of the arrangement of $(\reg_k-1)$ distinct parallel hyperplanes $H_k^p=\{\M x\inR^d\colon \M w_k^T\M x=b_k^p\}$, $p=1,\ldots,(\reg_k-1)$, and such that the hyperplanes are in general position when only one per partition is selected. In such a way, the arrangement $\mathcal{A}(\partition_1,\ldots,\partition_N)$ has exactly $\ub^d(\reg_1,\ldots,\reg_N)$ convex regions (see proof of Theorem \ref{th:upperbound}). Second, for each partition, we consider a CPWL function $\varphi_k\colon\R\rightarrow\R$ with knots $(b_k^p)_{p=1}^{\reg_k-1}$ and $\reg_k$ distinct affine pieces $(\varphi_k^p)_{p=1}^{\reg_k}$. In the sequel, the affine pieces are written $\varphi_k^p\colon x\rightarrow a_k^p x + c_k^p$. The function $f_k\colon \M x\mapsto \varphi(\M w_k^T\M x)$ has exactly $n_k$ linear convex regions and $\partition_k$ is a linear convex partition of it. The construction implies that $\mathcal{A}(\partition_1,\ldots,\partition_N)$ is a linear convex partition of both $(f_1+\cdots+f_N)$ and $(f_1,\ldots,f_N)$. Because the affine pieces of each $\varphi_k$ are distinct, the vector-valued function $(f_1,\ldots,f_N)$ will agree with distinct affine pieces on each region of $\mathcal{A}(\partition_1,\ldots,\partition_N)$, which proves that this partition has the minimal number of linear convex regions. This yields CPWL functions such that $\reg_{(f_1,\ldots,f_N)}= \ub^d(\reg_1,\ldots,\reg_N)$. On the contrary, in the case of the sum $(f_1+\cdots+f_N)$, there is on the contrary no guarantee that $\mathcal{A}(\partition_1,\ldots,\partition_N)$ is a partition with the minimal number of linear convex regions. To ensure that the regions of this partition have different affine pieces, it is sufficient to choose the pieces $(\varphi_k^p)$ such that $\sum_{k=1}^N \varphi_k^{p^1_k}(\M w_k^T\cdot)\neq \sum_{k=1}^N \varphi_k^{p^2_k}(\M w_k^T\cdot)$ for any $1\leq p_k^{1},p_k^{2} \leq \reg_k$ and $(p_1^1,\ldots,p_N^1)\neq (p_1^2,\ldots,p_N^2)$. An explicit choice is $a_k^p=p m^{k-1}$ with $m=\max(\reg_k)$. The biases $b_k^p$ are then set such that $\varphi_k^p$ is continuous. In such a way, the slope of $\sum_{k=1}^N \varphi_k^{p_k}(\M w_k^T\cdot)$ is $\sum_{k=1}^N p_k m^k$. This number can be represented in base $m$ as ``$(p_N\cdots p_1)_m$'', which shows that it is uniquely related to the choice of indices $(p_k)$. Although this choice seems very specific, a random choice of the slopes would also satisfy the condition almost surely. We have therefore found a collection of CPWL functions whose sum has exactly $\ub^d(\reg_1,\ldots,\reg_N)$ linear convex regions.
\end{proof}

\subsection{Compositional Bounds}

\label{ap:compositionalbounds}

\begin{proof}{\bf of Theorem \ref{th:boundcomposition}}
  We use the notation $m_\ell = \min(d_1,\ldots,d_\ell)$ and $\V F_\ell = \V f_{\ell}\circ \cdots \circ \V f_1$.
  First, we prove by induction the validity of the proposed upper bound. The initial step is given by Proposition \ref{pr:sumandvectorization}. Now suppose that the result holds for $\V F_{\ell-1}$ with $\ell-1>0$. Let $\Omega$ be a linear convex region of $\V F_{\ell-1}$ and let $\V g_{\Omega}$ be the corresponding affine function. The affine space $\V g_{\Omega} (\mathbb{R}^{d_1})\subset \mathbb{R}^{d_{\ell}}$ is of dimension at most $\min(d_1,\cdots,d_{\ell})$ (Lemma \ref{lm:dim}). Each linear convex partition $\partition_{{\ell,k}}$ of the components of $\V f_\ell$ yields a convex partition $\partition'_{\ell,k}$ of the affine subspace $\V g_{\Omega}(\mathbb{R}^{d_1})$ with no more than $\reg_{{\ell,k}}$ regions on which $f_{\ell,k}$ is affine (Lemma \ref{lm:subpartition}).
  The arrangement of the partitions $\partition'_{{\ell,1}},\ldots,\partition'_{{\ell,d_{\ell+1}}}$ results in a convex partition of $\V g_{\Omega} (\mathbb{R}^{d_1})$ with no more than $\ub^{m_\ell}(\reg_{\ell,1},\ldots,\reg_{\ell,{d_\ell+1}})$ regions (Theorem \ref{th:upperbound}). Lemma \ref{lm:inverseimagepartition} shows that $\V g_{\Omega}^{-1}(\mathcal{A}(\partition'_{{\ell,1}},\ldots,\partition'_{{\ell,d_{\ell+1}}}))$ is a convex partition of $\mathbb{R}^{d_1}$ with $\V F_\ell$ affine on each of its sets. In short, each linear convex region of $\V F_{\ell-1}$ is partitioned into no more than $\ub^{m_\ell}(\reg_{\ell,1},\ldots,\reg_{\ell,{d_{\ell+1}}})$ linear convex regions, which concludes the first part of the proof.

  Second, we propose a construction inspired from \cite{montufar2014number} to derive the lower bound given on the maximal number of regions. Let the sawtooth function $\mathrm{sw}_p$ of order $p$ be the unique $1$D CPWL function with knots located at $k/p$ for $k=1,\ldots,(p-1)$ that satisfies $\mathrm{sw}_p(k/p) = \frac{1}{2}(1-(-1)^k)$ for $k=0,\ldots,p$. The key properties of the sawtooth function of order $p$ that will prove useful in the sequel are
  \begin{itemize}
    \item  it has $p$ projection regions that are also convex linear regions;
    \item it can be decomposed as
    \begin{equation}
      \mathrm{sw}_p\colon x\mapsto \sum_{k=1}^p \varphi_{k,p},
    \end{equation}
    where $\varphi_{k,p}=p(x+2(-1)^p|x-k/p|)$ is a CPWL function with 2 projection regions;
    \item the composition of sawtooth functions is a sawtooth function whose order is the product of the orders of the composed functions, as in
    \begin{equation}
      \mathrm{sw}_p\circ\mathrm{sw}_q=\mathrm{sw}_{pq},
    \end{equation}
    for $p,q\in\N$.
  \end{itemize}
  The strategy is now to build a CPWL NN which mimics a given NN with independent sawtooth components.
  Let $\M e_{d,k}$ be the $k$th element of the canonical basis of $\R^d$, $d^*=\min (d_\ell)$ and $\tau_\ell\colon \{1,\ldots,d_{\ell+1}\}\rightarrow \{1,\ldots,d^*\}$ for $\ell=1,\ldots,L$. Consider the dimension-reduction linear operator $\V u_\ell\colon\R^{d_\ell}\rightarrow \R^{d^*}$ associated to $\tau_{\ell-1}$, which is defined on the canonical basis by
  \begin{equation}
    \V u_\ell\colon\M e_{d_\ell,k} \mapsto \M e_{d^*,\tau_{\ell-1}(k)},
  \end{equation}
  for $\ell=2,\ldots,L$ and 
  \begin{equation}
    \V u_1(\M e_{d_\ell,k})= \begin{cases}
      e_{d^*,k},& k\leq d^*\\
      0,& \text{ otherwise}.
    \end{cases}
    \end{equation}
    Similarly, let $\V v_\ell\colon \R^{d^*}\rightarrow \R^{d_{\ell+1}}$ be the dimension-augmentation linear operator
    \begin{equation}
      \V v_\ell\colon\M e_{d^*,k} \mapsto \sum_{q\in \tau_{\ell}^{-1}(\{k\})}\M e_{d_{\ell+1},q},
    \end{equation}
    for $\ell=1,\ldots,L$.

  We now define the nonlinear pointwise 
  function $\V \phi_{\ell}\colon \R^{d_{\ell+1}}\rightarrow\R^{d_{\ell+1}}$. For $r\in\{1,\ldots,d^*\}$ 
  let $p_{\ell,r} = \sum_{k\in \tau_{\ell}^{-1}(\{r\})} \reg_{\ell,k}$ and $\{J_{\ell,\tau_\ell,i}\}_{i=1}^{|\tau_{\ell}^{-1}(\{r\})|}$ be a partition of the set 
  $\{1,\ldots,p_{\ell,r}\}$, where the cardinality of the subsets is in one-to-one correspondence with $\{\reg_{\ell,q}\}_{q\in \tau_{\ell}^{-1}(\{r\})}$. In this way, we assign to each $k\in\{1,\ldots,{d_\ell+1}\}$
   a set of indices $J_{\ell,\tau_\ell,i_k}$ that allows us to define the $k$th component of $\V \phi_\ell$ as
  \begin{align}
    \phi_{\ell,k} = \sum_{j\in J_{\ell,\tau_\ell,i_k}} \varphi_{|\tau_{\ell}^{-1}(\{\tau_\ell (k)\})|,j}.
  \end{align}
  This ensures that
  \begin{align}
    \sum_{k\in \tau_{\ell}^{-1}(\{r\})} \phi_{\ell,k}= \mathrm{sw}_{p_{\ell,r}}.
  \end{align}
From the pointwise property of $\V \phi$, we deduce that, for any $t_1,\ldots,t_{d^*}\inR$,
 \begin{align}
  \nonumber
  (\V u_{\ell+1}\circ \V \phi \circ \V v_{\ell})\left(\sum_{r=1}^{d^*} t_r \M e_{d^*,r}\right) &= (\V u_{\ell+1}\circ\V \phi)\left(\sum_{r=1}^{d^*} \sum_{k\in \tau_{\ell}^{-1}(\{r\})}t_r\M e_{d_\ell,k}\right)\\
  \nonumber
  &=  \V u_{\ell+1}\left(\sum_{r=1}^{d^*}\sum_{k\in \tau_{\ell}^{-1}(\{r\})} \phi_{\ell,k}(t_r)\M e_{d_\ell,k}\right)\\
  \nonumber
  &= \sum_{r=1}^{d^*}\sum_{k\in \tau_{\ell}^{-1}(\{r\})} \phi_{\ell,k}(t_r) \V u_{\ell+1}(\M e_{d_\ell,k})\\
  \nonumber
  &= \sum_{r=1}^{d^*}\sum_{k\in \tau_{\ell}^{-1}(\{r\})} \phi_{\ell,k}(t_r) \M e_{d^*,\tau_\ell(k)}\\
  \nonumber
  &= \sum_{r=1}^{d^*}\left(\sum_{k\in \tau_{\ell}^{-1}(\{r\})} \phi_{\ell,k}(t_r)\right) \M e_{d^*,r}\\
  &= \sum_{r=1}^{d^*}\mathrm{sw}_{p_{\ell,r}}(t_r) \M e_{d^*,r},
\end{align}
which means that $\V u_{\ell+1}\circ\V \phi\circ\V v_{\ell}$ is a pointwise multivariate function with 1D sawtooth components of order $p_{\ell,r}$ for $r=1,\ldots,d^*$. We denote it by ${\mathrm{\mathbf{sw}}_{\M p_\ell}}$ with $\M p_\ell=(p_{\ell,1},\ldots,p_{\ell,d^*})$. The function $\V f_\ell$ of the NN is chosen to be $\V f_\ell = \V \phi_\ell \circ \V v_\ell \circ \V u_{\ell}$. Each component $f_{\ell,k}$ can be written in the form of $f_{\ell,k}\colon \M x \mapsto \phi_{\ell,k}(\M w_{\ell,k}^T \M x)$ with $\M w_{\ell,k}=\sum_{q\in\tau_{\ell-1}^{-1}(\{\tau_{\ell}(k)\})} \M e_{d_\ell,q}$. This shows that $f_{\ell,k}$ has the same number of projection regions as $\phi_{\ell,k}$ ($\reg_{\ell,k}$) whenever $\M w_{\ell,k}\neq \M 0$.

All in all, we have that
\begin{align}
  \nonumber
  \V f_L\circ\V f_{L-1}\circ \cdots \circ \V f_2\circ \V f_1 &= (\V \phi_L \circ \V v_L \circ \V u_{L}) \circ (\V \phi_{L-1} \circ \V v_{L-1} \circ \V u_{L-1}) \circ \cdots \circ (\V \phi_{2} \circ \V v_{2} \circ \V u_{2})\circ (\V \phi_{1} \circ \V v_{1} \circ \V u_{1})\\
  \nonumber
  &= \V \phi_L \circ \V v_L \circ (\V u_{L} \circ \V \phi_{L-1} \circ \V v_{L-1}) \circ (\V u_{L-1} \circ \cdots \circ \V \phi_{2} \circ \V v_{2}) \circ (\V u_{2}\circ \V \phi_{1} \circ \V v_{1}) \circ \V u_{1})\\
  &= \V \phi_L \circ \V v_L \circ {\mathrm{\mathbf{sw}}_{\M p_{L-1}}} \circ \cdots \circ {\mathrm{\mathbf{sw}}_{\M p_{1}}} \circ \V u_{1}.
\end{align}
We now note that there are no fewer projection regions of $\V f_L\circ\V f_{L-1}\circ \cdots \circ \V f_2\circ \V f_1$ than the number of projection regions of $\V h=\V u_{L+1}\circ\V f_L\circ \cdots \circ \V f_1$ because $\V u_{L+1}$ is a linear mapping. In addition,
\begin{align}
  \nonumber
  \V u_{L+1}\circ\V f_L\circ \cdots \circ \V f_1 &= {\mathrm{\mathbf{sw}}_{\M p_{L}}} \circ \cdots \circ {\mathrm{\mathbf{sw}}_{\M p_{1}}} \circ \V u_{1}\\
  &= {\mathbf{sw}}_{\M q} \circ \V u_{1},
\end{align}
where $\M q=(q_1,\ldots,q_{d^*})$ and $q_r=\prod_{\ell=1}^L p_{\ell,r}$. The properties of the sawtooth functions and the special form of $\V u_{1}$ yields the projection regions for $\V h$ as
\begin{equation}
  \{\M x\inR^{d_1}\colon \text{for } r=1,\ldots,d^*, \begin{cases}-\infty < x_r \leq 1/q_r, & i_r = 0\\
    1-1/q_r \leq x_r < +\infty, & i_r = q_r-1\\
    i_r/q_r \leq x_r \leq (i_r+1)/q_r, & \text{otherwise}
  \end{cases} \},
\end{equation}
where $i_r=0,\ldots,(q_r-1)$ for $r= 1,\ldots,d^*$. In summary, the number of projection regions of the constructed CPWL NN is at least
\begin{equation}
  \prod_{r=1}^{d^*} q_r = \prod_{\ell=1}^L \prod_{r=1}^{d^*} p_{\ell,r} = \prod_{\ell=1}^L \prod_{r=1}^{d^*} \sum_{k\in \tau_{\ell}^{-1}(\{r\})} \reg_{\ell,k}.
\end{equation}
The conclusion is reached by noticing that the reasoning does not depend on any property of the mappings $\tau_\ell$: one can therefore pick the ones that yield the largest lower bound.

\end{proof}
\begin{proof}{\bf of Corollary \ref{cr:boundcompo}} To get the upper bound, we combine Theorem \ref{th:boundcomposition} and the simplified version of the bound given in Theorem \ref{th:upperbound} with the assumption that $W\geq d_{\mathrm{in}}\geq d^*\coloneqq\min(d_{\mathrm{in}},d_{\mathrm{out}})$. To get the lower bound, we have to compute \begin{equation}
    \alpha^{d^*}(\reg_{\ell,1},\ldots,\reg_{\ell,{d_{\ell+1}}}) = \max_{\tau \in \mathcal{T}_{d_{\ell}}}\prod_{r=1}^{d^*}\sum_{k\in\tau^{-1}(\{r\})}\reg_{\ell,k}.
  \end{equation}
  We lower-bound this quantity by selecting an arbitrary mapping $\tau\colon [W]\rightarrow [d^*]$ such that, for $r\in[d^*]$, the cardinality of $\tau^{-1}(\{r\})$ is at least $\floor{W/d^*}$. In this way, we obtain that
  \begin{equation}
    \alpha^{d^*}(\reg_{\ell,1},\ldots,\reg_{\ell,{d_{\ell+1}}})\geq (\kappa \floor{W/d^*})^{d^*}
  \end{equation}
  for $\ell=1,\ldots,L$ and reach the given lower bound.
  \end{proof}
  \begin{proof}{\bf of Corollary \ref{cr:boundcompoprojection}}
    The upper bound is a direct consequence of Proposition \ref{pr:projectionvsconvex}: the number of convex linear regions is never smaller than the number of projection regions. The CPWL NN built to provide the lower bound of Theorem \ref{th:boundcomposition} had exactly as many projection regions as convex linear regions, hence justifying the lower bound.
  \end{proof}

\section{\changer{Proofs for the Knot Density of the Sum, Vectorization, and Composition of CPWL Functions}}
\label{ap:density_bounds_operation}
\begin{proof}{\bf of Proposition \ref{pr:densitysumveccompo}}
  Consider two CPWL functions $\V f_1,\V f_2$ with characteristic functions $\V\varphi_1^{\V \curve}$ and $\V\varphi_2^{\V \curve}$ along ${\V \curve}$ and projection regions $(\Omega_k^1)_{k=1}^{K_1}$ and $(\Omega_k^2)_{k=1}^{K_2}$. Let $\V\varphi_{(1,2)}^{\V \curve}$ denote the characteristic function of $(\V f_1,\V f_2)$ and $\V\varphi_{1+2}^{\V \curve}$ the one of $\V f_1 + \V f_2$ along ${\V \curve}$. Consider a subset $R\subset S$ on which $\V\varphi_1^{\V \curve}$ and $\V\varphi_2^{\V \curve}$ are continuous. Following the definition of the characteristic function, any projection region $\Omega$ of $\V f_1$ or of $\V f_2$ either entirely contains $\V \curve(R)$ or does not intersect with it; otherwise, $\V\varphi_1^{\V \curve}$ or $\V\varphi_2^{\V \curve}$ would not be continuous on $R$. Any projection region $\Omega^{(1,2)}$ of $(\V f_1,\V f_2)$ is a nonempty intersection of the form $\Omega^1_p\cap\Omega^2_q, $ with $p\in[K_1]$ and $q\in[K_2]$. Since the regions $\Omega^1_p$ and $\Omega^2_q$ either entirely contain $\V \curve(R)$ or do not intersect with it, the same holds true for $\Omega^{(1,2)}$, which implies that $\V \varphi_{(1,2)}^{\V \curve}$ must be continuous on $R$. The same argument holds true for $\V \varphi_{(1+2)}^{\V \curve}$, the only difference being that the projection regions of $\V f_1+\V f_2$ are unions of nonempty subsets of the form $\Omega^1_p\cap\Omega^2_q$, which also has the same implications. So, on the one hand, we proved that, where $\V\varphi_1$ and $\V\varphi_2$ are continuous, $\V\varphi_{1+2}$ and $\V\varphi_{(1,2)}$ are also continuous. On the other hand, the number of points where either $\V\varphi_1$ or $\V\varphi_2$ is discontinuous is no greater than the number of points where $\V\varphi_1$ and $\V\varphi_2$ are discontinuous, which concludes the proof.
\end{proof}

\begin{proof}{\bf of Proposition \ref{pr:compolinear}}
  Consider the characteristic function $\V \varphi_{\V f_1}^{\V \curve}$ of $\V f_1$ on $\V\curve$ and $\V \varphi_{\V f_2}^{\V f_1\circ\V \curve}$ of $\V f_2$ on $\V f_1\circ\V \curve$. Moreover, consider a subset $R\subset S$ on which both $\V \varphi_{\V f_1}^{\V \curve}$ and $\V \varphi_{\V f_2}^{\V f_1\circ\V \curve}$ are continuous. The projection regions of $\V f_1$ and $\V f_2$ are denoted by $(\Omega_k^1)_{k=1}^{K_1}$ and $(\Omega_k^2)_{k=1}^{K_2}$. On any region $\Lambda_{p,q} = \Omega_p^1\cap \V f_1^{-1}(\Omega_q^2)$, the function $\V f_2\circ\V f_1$ is affine, meaning that each projection region of $\V f_2\circ\V f_1$ is a union of some of the regions $\Lambda_{p,q}$. Each region $\Omega_p^1$ either entirely contains the subset $\V\curve(R)$ or does not intersect with it; otherwise, $\V \varphi_{\V f_1}^{\V \curve}$ would not be continuous on $R$. We now remark that $\V \varphi^{\V f_1\circ\V\curve}_{\V f_2}(t)=(\mathbbm{1}_{\Omega_1^2}(\V f_1\circ \V\curve(t)),\ldots,\mathbbm{1}_{\Omega_{K_2}^2}(\V f_1\circ\V\curve(t))) = (\mathbbm{1}_{\V f_1^{-1}(\Omega_1^2)}(\V\curve(t)),\ldots,\mathbbm{1}_{\V f_1^{-1}(\Omega_{K_2}^2)}(\V\curve(t)))$, which means that each region $\V f_1^{-1}(\Omega_q^2)$ either entirely contains the subset $\V\curve(R)$ or does not intersect with it since, otherwise, $\V \varphi_{\V f_2}^{\V f_1\circ\V \curve}$ would not be continuous on $R$. We therefore have that each region $\Lambda_{p,q}$ either entirely contains the subset $\V\curve(R)$ or does not intersect with it. Consequently, the same holds true for union of regions $\Lambda_{p,q}$ and, as a result, for the projection regions of $\V f_2\circ\V f_1$. This shows that, where $\V \varphi_{\V f_1}^{\V \curve}$ and $\V \varphi_{\V f_2}^{\V f_1\circ\V \curve}$ are continuous, $\V \varphi_{\V f_2\circ\V f_1}^{\V \curve}$ is also continuous. In short, the number of points of discontinuities of $\V \varphi_{\V f_2\circ\V f_1}^{\V \curve}$ is no greater than the the number of points of discontinuity of either $\V \varphi_{\V f_1}^{\V \curve}$ or $\V \varphi_{\V f_2}^{\V f_1\circ\V \curve}$, which concludes the proof.
\end{proof}

\section{Proof of the Bounds on the Knot Density of Classical CPWL Components}
\label{ap:densitybounds}
\begin{proof}{\bf of Proposition \ref{pr:knotdensityReLU}}
  First, we prove the result when $\V \curve$ parameterizes a linear segment. Let $\M x_0,\M u\inR^d$ with $\|\M u\|_2 = 1$ and $\V \curve\colon t\mapsto \M x_0 + t \M u$ for $t\in S$, where $S=[0,|S|]\subset \R$ is a segment. The first step is to compute the probability $\mathbb{P}(\mathrm{kt}_f^{\V \curve}=1)$ that $f$ has a knot along $\V \curve$. The hyperplane $\{\M x\inR^d\colon \V w^T \M x + b = 0\}$ intersects the line $\{\M x_0 + t \M u\colon t\inR\}$ for $t_0$ such that $\V w^T (\M x_0 + t_0\M u) + b = 0$ or, equivalently, $b = (-\V w^T (\M x_0 + t_0\M u))$. In order to have a knot along $\V \curve_{|S}$, $t_0$  has to lie in $S$. For a given $\V w$, this implies that $b$ should be in an interval of length $|S||\V w^T\M u|$, more precisely $[-\V w^T\M x_0,|S|\V w^T\M u-\V w^T\M x_0]$ if $\V w^T\M u<0$ and $[|S|\V w^T\M u-\V w^T\M x_0,-\V w^T\M x_0]$ otherwise. Therefore,
  $\mathbb{P}(\mathrm{kt}_f^{\V \curve}=1|\V w) \leq \sup_{t\inR}\rho_b(t)|S||\V w^T\M u|$. From the independence of the random variables and from the fact that $\mathrm{kt}_f^{\V \curve}=0$ or $\mathrm{kt}_f^{\V \curve}=1$ almost surely, we infer that $\mathbb{E}[\density_f^{\V \curve_{|S}}] \leq \sup_{t\inR}\rho_b(t)\mathbb{E}[|\V w^T\M u|]\leq \sup_{t\inR}\rho_b(t) \sqrt{\mathbb{E}[|\V w^T\M u|^2]} = \sup_{t\inR}\rho_b(t) \sqrt{\M u^T \mathbb{E}[\V w \V w^T] \M u }\leq \sup_{t\inR}\rho_b(t) \sqrt{\mathbb{E}[w^2]}$. In the last step, the assumption that the random variables $w_k$ are i.i.d. has allowed us to infer that $\mathbb{E}[\V w \V w^T]=\mathbb{E}[w^2]\M I$, where $\M I\inR^{d\times d}$ is the identity matrix.
  
  If $b$ is normally distributed with standard deviation $\sigma_b$, then $\sup_{t\inR}\rho_b(t) = (\sigma_b \sqrt{2\pi})^{-1}$. In addition, suppose that the components $w_k$ are independent and normally distributed with standard deviation $\sigma_w$. The random variable $\V w^T \M u$ is also normally distributed with standard deviation $\sigma_w$ (since $\|\M u\|_2 = 1$). We can now compute explicitly $\mathbb{E}[|\V w^T\M u|]=\sigma_w\sqrt{2}/\sqrt{\pi}$ based on the properties of half-normal distributions.
  
  The result is extended to any polygonal chain through the linearity of the expectation operator and by application of the result to the finitely many pieces of the polygonal chain.
  \end{proof}
\begin{proof}{\bf of Proposition \ref{pr:knotdensityMaxout}}
  A knot of $f$ along a line $\V \curve$ must lie on a hyperplane $H_{p,q} = \{\M x\colon (\V w_p - \V w_q)^T \M x + (b_p-b_q)=0\}$ with $1\leq p< q \leq K$, since elsewhere the Maxout unit is affine. Therefore, the expected knot density is bounded as
  \begin{align}
    \nonumber
    \mathbb{E}\left[\density_{\mathrm{Maxout}}^{\V \curve}\right] &\leq \frac{1}{|S|} \mathbb{E}\left[\sum_{1\leq p < q \leq K} (\V \curve(S)\cap H_{p,q}\neq \emptyset)\right]\\
    \nonumber
    &= \frac{1}{|S|} \sum_{1\leq p < q \leq K} \mathbb{E}\left[(\V \curve(S)\cap H_{p,q}\neq \emptyset)\right]\\
    \nonumber
    &\leq \frac{1}{|S|} \sum_{1\leq p < q \leq K} \sqrt{\mathbb{E}\left[(w_p-w_q)^2\right]} \sup_{t\inR}\rho_b(t)\\
    \nonumber
    &= \frac{1}{|S|} \sum_{1\leq p < q \leq K} \sqrt{2}\sigma_w \sup_{t\inR}\rho_b(t)\\
    &= \frac{1}{|S|} \sqrt{2}{K\choose 2}\sigma_w\sup_{t\inR}\rho_b(t),
  \end{align}
 where we have taken advantage of the results derived in the proof of Proposition \ref{pr:knotdensityReLU} to bound the probability that a randomly generated hyperplane intersects a segment of length $|S|$, along with the independence of the random variables. We also notate $(A\neq \emptyset)$ to encode the variable that takes the value 0 if $A=\emptyset$ and $1$ otherwise. When the random variables are normally distributed, the reasoning is similar to the one in the proof of Proposition \ref{pr:knotdensityReLU}.
\end{proof}
\begin{proof}{\bf of Proposition \ref{pr:knotdensityGS}}
  A knot of $f$ along a line $\V \curve$ must lie on a hyperplane $H_{p,q} = \{\M x\colon (\V w_p - \V w_q)^T \M x + (b_p-b_q)=0\}$, where $1\leq p < q \leq K$ and where $p,q$ belong to the same sorting group, since elsewhere the GroupSort layer is affine. One can now follow the same steps as those in the proof of Proposition \ref{pr:knotdensityMaxout} with $n_g {g_s \choose 2}= n_g g_s(g_s-1)/2= d (g_s-1)/2$ hyperplanes.
\end{proof}

\section{\changer{Proofs of the Bounds on the Expected Knot Density of CPWL NNs}}
\label{ap:density_cpwl_nn}

\begin{proof}{\bf of Lemma \ref{lm:compolength}}
  In what follows, the technical developments originate from the fact that $\V f_{\V \theta}$ is not differentiable everywhere.
The function $\V f_{\V \theta}\circ\V\curve$ is the composition of two CPWL functions, hence it is CPWL and, therefore, differentiable for almost every $t\in S$. Note, however, that we cannot assert that the Jacobian of $\V f_{\V \theta}$ is well defined at $\V \curve(t)$ for almost every $t\in S$. Indeed, whenever $\V\curve$ follows the boundary of two projection regions of $\V f_{\V \theta}$, the Jacobian of $\V f_{\V \theta}$ along $\V \curve$ becomes ill-posed. This is why the notion of directional derivative is better suited. The characteristic function $\V\varphi_{\V f_{\V \theta}}^{\V \curve}$ of $\V f_{\V \theta}$ along $\V\curve$ is piecewise-constant on $S$: We can partition $S$ into finitely many convex regions where $\V\varphi_{\V f_{\V \theta}}^{\V \curve}$ is constant. Let $P\subset S$ denote one of these regions and let $Q\subset S$ be a linear convex region of $\V\curve$. Following the definition of the characteristic function, there exists a projection region $\Omega$ of $\V f_{\V \theta}$ such that $\V \curve(\mathrm{int}(P\cap Q))$ either lies entirely in the interior of $\Omega$, or entirely on its boundary.
 In the first case, $\V f_{\V \theta}$ is differentiable in $\V \curve(\mathrm{int}(P\cap Q))$ and we have that $(\V f_{\V \theta}\circ\V\curve)'(t) = J_{\V f_{\V \theta}}(\V \curve(t))\V \curve'(t)= D_{\V \curve'(t)}\{\V f_{\V \theta}\}(\V\curve(t))$ for $t\in\mathrm{int}(P\cap Q)$. In the second case,  $\V\curve$ is differentiable on $\mathrm{int}(P\cap Q)$ as well, but the Jacobian of $\V f_{\V \theta}$ is undefined. Fortunately, the directional derivative of $\V f_{\V \theta}$ is well defined along $\V \curve(t)$ since, for any $t\in\mathrm{int}(P\cap Q)$, there exists $\epsilon>0$ such that $\tau\mapsto\V f_{\theta}(\V\curve(t)+\tau \V\curve'(t))$ is affine on $(-\epsilon,\epsilon)$. All in all, the relation $(\V f_{\V \theta}\circ\V\curve)'(t)=D_{\V \curve'(t)}\{\V f_{\V \theta}\}(\V\curve(t))$ is well defined for any $t\in \mathrm{int}(P\cap Q)$ and, more generally, for almost any $t\in S$ because of the properties of $P$ and $Q$. We can now write that
  \begin{align}
    \mathbb{E}\left[\length(\V f_{\V \theta}\circ\V\curve)\right] &=\mathbb{E}\left[\int_{t\in S} \left\|(\V f_{\V \theta}\circ\V\curve)'(t)\right\|_2 \dint t\right] \nonumber\\
    &=\mathbb{E}\left[\int_{t\in S} \left\|D_{\V \curve'(t)}\{\V f_{\V \theta}\}(\V\curve(t))\right\|_2 \dint t\right]\nonumber\\
    &=\int_{t\in S} \mathbb{E}\left[\left\|D_{\V \curve'(t)}\{\V f_{\V \theta}\}(\V\curve(t))\right\|_2 \right]\dint t\nonumber\\
  &\leq \sup_{\substack{\M x,\M u \inR^d\nonumber\\ \|\M u\|_2 = 1}}\mathbb{E}[\|D_{\M u}\{\V f_{\V \theta}\}(\M x)\|_2] \int_{t\in S} \left\|\V \curve'(t)\right\|_2\dint t\nonumber\\
  &= \length(\V \curve) \sup_{\substack{\M x,\M u \inR^d\\ \|\M u\|_2 = 1}}\mathbb{E}[\|D_{\M u}\{\V f_{\V \theta}\}(\M x)\|_2],
  \end{align}
  where we have used Tonelli's theorem to interchange the expectation and the integral.
\end{proof}

\begin{proof}{\bf of Theorem \ref{th:boundstochastic}}
  Let $\V F_\ell = \V f_{\V \theta_\ell}\circ\cdots\circ\V f_{\V \theta_1}$. With Lemma \ref{pr:compolinear}, we have that
  \begin{align}
    \mathbb{E}\left[\mathrm{kt}_{\V F_\ell}^{\V \curve}\right] = \mathbb{E}\left[\mathrm{kt}_{\V f_{\V \theta_\ell}\circ \V F_{L-1}}^{\V \curve}\right] \leq \mathbb{E}\left[\mathrm{kt}_{\V f_{\V \theta_\ell}}^{\V F_{\ell-1} \circ \V \curve}\right] + \mathbb{E}\left[\mathrm{kt}_{\V F_{\ell-1}}^{\V \curve}\right].
  \end{align}
  We now apply the law of the iterated expectation to obtain that
  \begin{align}
    \nonumber
    \mathbb{E}\left[\mathrm{kt}_{\V f_{\V \theta_\ell}}^{\V F_{\ell-1} \circ \V \curve}\right] &= \mathbb{E}_{\V \theta_1,\ldots,\V \theta_{\ell-1}}\left[\mathbb{E}_{\V \theta_{\ell}}\left[\mathrm{kt}_{\V f_{\V \theta_\ell}}^{\V F_{\ell-1} \circ \V \curve}|\V \theta_1,\ldots,\V \theta_{\ell-1}\right]\right]\\
    &\leq \mathbb{E}_{\V \theta_1,\ldots,\V \theta_{\ell-1}}\left[d \density_0 \length(\V F_{\ell-1} \circ \V \curve)\right],
  \end{align}
  where the inequality follows from the first assumption of the theorem, the application of Proposition \ref{pr:densitysumveccompo}, and requires the independence of the random variables. We can now apply Lemma \ref{lm:compolength} recursively to $\V F_\ell$ and invoke the second assumption of the theorem to infer that
  \begin{align}
    \mathbb{E}\left[\mathrm{kt}_{\V f_{\V \theta_\ell}}^{\V F_{\ell-1} \circ \V \curve}\right] &\leq d \density_0 \length(\V \curve)D_0^{\ell-1}.
  \end{align}
  All in all, we just proved that 
\begin{align}
  \mathbb{E}\left[\mathrm{kt}_{\V F_\ell}^{\V \curve}\right] \leq \mathbb{E}\left[\mathrm{kt}_{\V F_{\ell-1}}^{\V \curve}\right] + d \density_0 \length(\V \curve)D_0^{\ell-1},
\end{align}
which reads in term of linear densities as
\begin{align}
  \label{eq:recrelation}
  \mathbb{E}\left[\density_{\V F_\ell}^{\V \curve}\right] \leq \mathbb{E}\left[\density_{\V F_{\ell-1}}^{\V \curve}\right] + d \density_0 D_0^{\ell-1}.
\end{align}
This recurrence relation directly yields the announced bound.
\end{proof}
\begin{proof}{\bf of Corollary \ref{cr:boundstochastic}}
  The proof is similar to the proof of Theorem \ref{th:boundcomposition} except that, with the different second assumption, the quantity $\mathbb{E}_{\V \theta_1,\ldots,\V \theta_{\ell-1}}[\length(\V F_{\ell-1} \circ \V \curve)]$ can be bounded by $D_0\length(\V \curve)$ (Lemma \ref{lm:compolength}). In the end the recurrence relation \eqref{eq:recrelation} is changed into
  \begin{align}
    \label{eq:recrelation2}
    \mathbb{E}\left[\density_{\V F_\ell}^{\V \curve}\right] \leq \mathbb{E}\left[\density_{\V F_{\ell-1}}^{\V \curve}\right] + D_0 \density_0 W.
  \end{align}
\end{proof}

\end{document}